\documentclass[11pt]{article}
\usepackage{amssymb}
\usepackage{algorithm}
\usepackage{algpseudocode}
\usepackage{subcaption}
\usepackage[absolute,overlay]
{textpos}
\usepackage{mathrsfs}
\usepackage{amsmath}
\usepackage{booktabs}
\usepackage{multirow}
\usepackage[table]{xcolor}
\usepackage{caption}
\usepackage{makecell}
\usepackage{enumitem}
\usepackage[most]{tcolorbox}

\usepackage[final]{acl}
\usepackage{times}
\usepackage{latexsym}

\usepackage[T1]{fontenc}

\usepackage[utf8]{inputenc}

\usepackage{microtype}

\usepackage{inconsolata}

\usepackage{graphicx}

\title{\textsc{CR-Seg}: Attention-Guided and CoT-Enhanced Coarse-to-Refined Reasoning Segmentation
}

\author{
\textbf{Yifan Cao},
\textbf{Xiaocui Yang\thanks{Corresponding author.}},
\textbf{Faxian Wan},
\textbf{Shi Feng},
\textbf{Daling Wang},
\textbf{Yifei Zhang} \\
School of Computer Science and Engineering, Northeastern University
Shenyang 110819, China \\
\texttt{\{caoyifan, wanfaxian\}@mails.neu.edu.cn} \\
\texttt{\{yangxiaocui, fengshi, wangdaling, zhangyifei\}@cse.neu.edu.cn} \\
}

\begin{document}
\maketitle
\begin{abstract}
Reasoning segmentation aims to segment target objects described by complex language through joint visual-textual reasoning. Existing methods typically rely on either learned semantic tokens to bridge Multimodal Large Language Models (MLLMs) and segmentation models, suffering from difficult cross-modal alignment, or explicit spatial prompts such as bounding boxes, which may lose holistic response semantics.
To address these limitations, we propose Attention-Guided and CoT-Enhanced \textbf{C}oarse-to-\textbf{R}efined Reasoning \textbf{Seg}mentation, termed \textsc{CR-Seg}, a two-stage framework for coarse-to-refined reasoning segmentation. Specifically, we design an Extract Attention Maps and Points (EAP) module to extract attention maps for coarse target localization and select informative points, both of which are fed into SAM for mask refinement. 
To alleviate reasoning--answer inconsistency, we further introduce Global-to-Local Chain-of-Thought (GLCoT), which guides the model to reason progressively from global scene context to local target details. Extensive experiments on reasoning segmentation benchmarks demonstrate the effectiveness of \textsc{CR-Seg}. 

\end{abstract}

\begin{figure}[t]
  \includegraphics[width=\columnwidth]{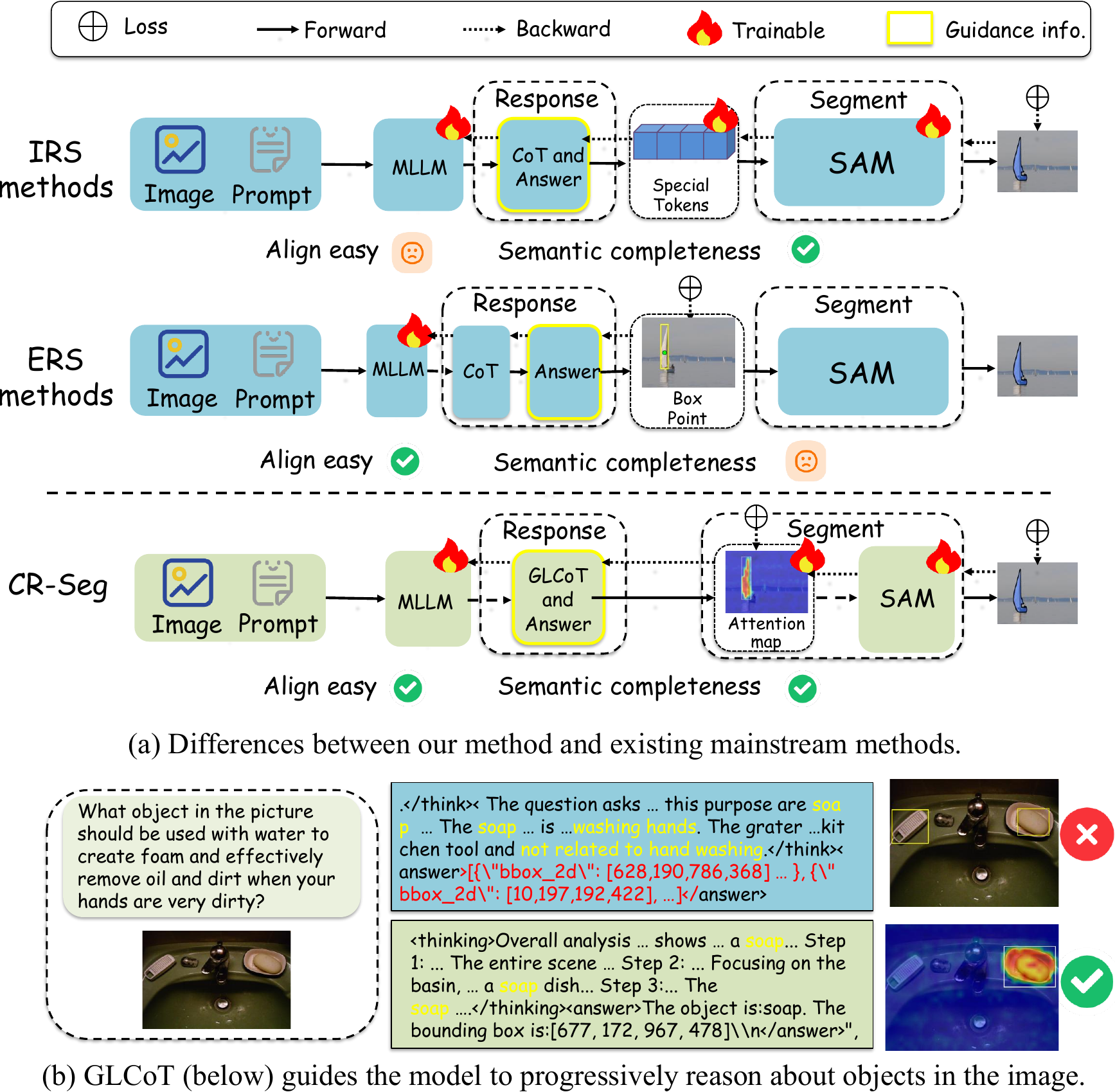}
  \caption{Difference between \textsc{CR-Seg} and existing methods.}
  \label{fig:intro}
\end{figure}

\section{Introduction}

Reasoning segmentation \citep{lai2024lisa} extends referring segmentation \citep{kazemzadeh2014referitgame,Rasheed_2024_CVPR,zhang2024omg} from explicit referring expressions to complex textual descriptions, requiring models to jointly understand visual--textual inputs and perform fine-grained reasoning. With Multimodal Large Language Models (MLLMs) \citep{NEURIPS2023_6dcf277e,bai2025qwen25vltechnicalreport} showing strong multimodal understanding and the Segment Anything Model (SAM) family \citep{Kirillov_2023_ICCV,ravi2025sam,carion2025sam3segmentconcepts} providing accurate mask generation, integrating MLLMs with SAM has become a widely adopted paradigm \citep{Wu_2024_CVPR,liu2025segzeroreasoningchainguidedsegmentation}.

As shown in Figure~\ref{fig:intro}(a), existing methods for bridging MLLMs and segmentation models can be roughly grouped into two paradigms, i.e., Internal-Representation-Based Reasoning Segmentation (IRS) and Explicit-Prompt-Based Reasoning Segmentation (ERS). IRS methods use hidden states of MLLM response tokens as the interface to segmentation models \citep{Xia_2024_CVPR,lai2024lisa}. These hidden states can aggregate holistic response semantics \citep{qian2024reasoning,DBLP:conf/aaai/ZhuOZCHSRCYLW26}, but they lack explicit spatial supervision and are optimized mainly through the final mask loss, making cross-modal alignment difficult. ERS methods \citep{liu2025segzeroreasoningchainguidedsegmentation,liu2026visionreasonerunifiedreasoningintegratedvisual} instead train MLLMs to produce explicit grounding cues, such as bounding boxes, which are then fed into segmentation models. Although this avoids latent alignment training, it compresses the full reasoning process into sparse spatial prompts. As a result, CoT reasoning does not directly guide mask generation, and segmentation quality becomes highly dependent on grounding accuracy. In practice, we observe that existing models are prone to reasoning--answer inconsistency. When distractors appear in later reasoning steps, the model may produce incorrect grounding predictions and thus erroneous masks, as illustrated in Figure~\ref{fig:intro}(b).

\begin{figure}[t]
  \centering
    \IfFileExists{image/image2_2.pdf}{%
      \includegraphics[width=\linewidth]{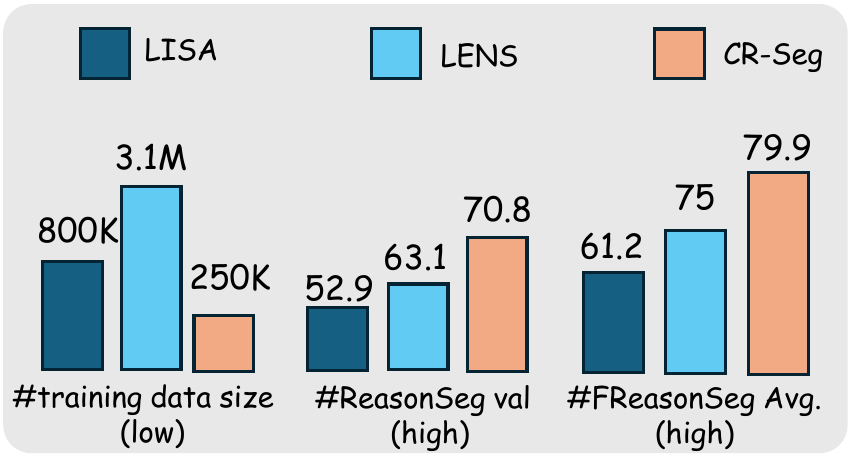}%
    }{%
      \framebox[\linewidth]{\rule{0pt}{180pt} figure1_concept.pdf}%
    }
    \vspace{-13pt} 
    \caption{Compared to other IRS methods, such as LISA \citep{lai2024lisa} and LENS \citep{DBLP:conf/aaai/ZhuOZCHSRCYLW26}, CR-Seg (ours) achieves better performance on ReasonSeg \citep{lai2024lisa} and FReasonSeg with less aligned training data($\text{epochs} \times \text{dataset size}$).}
  \label{fig:2_2}
\end{figure}

These limitations suggest the need for a spatial--semantic bridge that is grounded in the MLLM response, explicitly supervisable, and compatible with SAM refinement. Recent studies on visual grounding \citep{Kang_2025_CVPR,zhou2026guiaimaaligningintrinsicmultimodal} show that attention maps encode token-level spatial semantics. Compared with IRS hidden states, attention maps expose dense spatial distributions that can be supervised at the mask level; compared with ERS prompts, they preserve richer response semantics instead of relying on a single discrete grounding output. Therefore, attention-derived priors can reduce the alignment difficulty of IRS while alleviating the semantic compression problem of ERS.
Based on this insight, we propose Attention-Guided and CoT-Enhanced \textbf{C}oarse-to-\textbf{R}efined Reasoning \textbf{Seg}mentation, termed \textsc{CR-Seg}. As illustrated in Figure~\ref{fig:intro}(a), \textsc{CR-Seg} is a two-stage framework that uses MLLM attention maps as coarse segmentation priors and refines them into final masks with the Segment Anything Model (SAM). We propose an Extract Attention Maps and Points (EAP) module to extract attention maps and derive informative point prompts, providing lightweight spatial-semantic priors for target localization. In the first stage, these attention-derived priors are aligned with segmentation masks, allowing SAM to focus on boundary refinement. In the second stage, we further introduce Global-to-Local Chain-of-Thought (GLCoT), a three-step reasoning strategy that guides the model to survey the global scene, identify candidate objects, and locate the target by excluding distractors. To evaluate fine-grained discrimination among same-category objects, we further construct FReasonSeg as an auxiliary benchmark. As shown in Figure~\ref{fig:2_2}, \textsc{CR-Seg} achieves strong performance with reduced alignment overhead.
Our main contributions are summarized as follows:
\begin{itemize}[leftmargin=*, itemsep=1pt, topsep=1pt, parsep=0pt]
\item We propose \textsc{CR-Seg}, an attention-guided coarse-to-refined reasoning segmentation framework that transfers MLLM spatial--semantic knowledge to SAM through mask-supervised attention priors, preserving holistic response semantics while reducing alignment difficulty.
\item We introduce GLCoT, a global-to-local reasoning strategy that filters distractors before final target localization, mitigating reasoning--answer inconsistency and improving segmentation accuracy.
\item We construct FReasonSeg, an auxiliary benchmark for fine-grained target discrimination. Extensive experiments demonstrate the effectiveness and robustness of \textsc{CR-Seg}.
\end{itemize}

\section{Related Work}
\subsection{Internal-Representation-Based Reasoning Segmentation (IRS)}
IRS methods, such as LISA \citep{lai2024lisa}, GSVA \citep{Xia_2024_CVPR}, and PixeLM \citep{ren2024pixellm}, use semantic tokens, e.g., [SEG] tokens, to guide segmentation. Prior work \citep{qian2024reasoning} has shown that [SEG] tokens can produce similarity maps aligned with target regions.  
Since [SEG] tokens cannot be directly supervised, they are optimized only through the final masks.
However, this indirect supervision typically incurs substantial alignment costs. Moreover, requiring MLLMs to explicitly emit [SEG] tokens constrains generation flexibility. LENS \citep{DBLP:conf/aaai/ZhuOZCHSRCYLW26} partially addresses this constraint via a dedicated context module that extracts semantic information without explicit [SEG] tokens, and shows that such semantic tokens can rectify certain grounding errors and recover correct masks. Nevertheless, it does not explicitly address reasoning--answer inconsistency and still requires considerable alignment training.
To this end, we use directly supervisable attention maps as the information bridge. This design preserves holistic semantic information from the full MLLM response while simplifying alignment. The resulting attention maps also serve as coarse segmentation priors, allowing the segmentation model to focus on fine-grained boundary refinement.

\subsection{Explicit-Prompt-Based Reasoning Segmentation (ERS)}
ERS methods decouple reasoning and segmentation into two independent stages, thereby avoiding complex alignment training. Representative works, such as SegZero \citep{liu2025segzeroreasoningchainguidedsegmentation} and VisionReasoner \citep{liu2026visionreasonerunifiedreasoningintegratedvisual}, train MLLMs to produce accurate grounding outputs (e.g., bounding boxes or points), which are then passed to a standalone segmentation model. Building on this paradigm, Dr.~Seg \citep{sun2026dr}, PixelThink \citep{wang2025pixelthinkefficientchainofpixelreasoning}, and DR$^2$Seg \citep{he2026dr2segdecomposedtwostagerollouts} pursue further gains through advances in reinforcement learning algorithms, training pipelines, and reward function design. More recently, training-free multi-step reasoning segmentation models, exemplified by SAM3-Agent \citep{carion2025sam3segmentconcepts} and Evol-SAM3 \citep{ye2025evolvingtrainingzeroshotreasoning}, improve performance via multi-step reasoning strategies, albeit at the cost of higher inference latency.
However, a fundamental limitation of this paradigm,  is that only the final grounding outputs are utilized for segmentation, while the semantic information embedded in the CoT reasoning is discarded. Consequently, any errors in grounding can directly propagate to the final segmentation mask.

\begin{figure*}[t]
  \centering
    \IfFileExists{image/image3.pdf}{%
      \includegraphics[width=\linewidth]{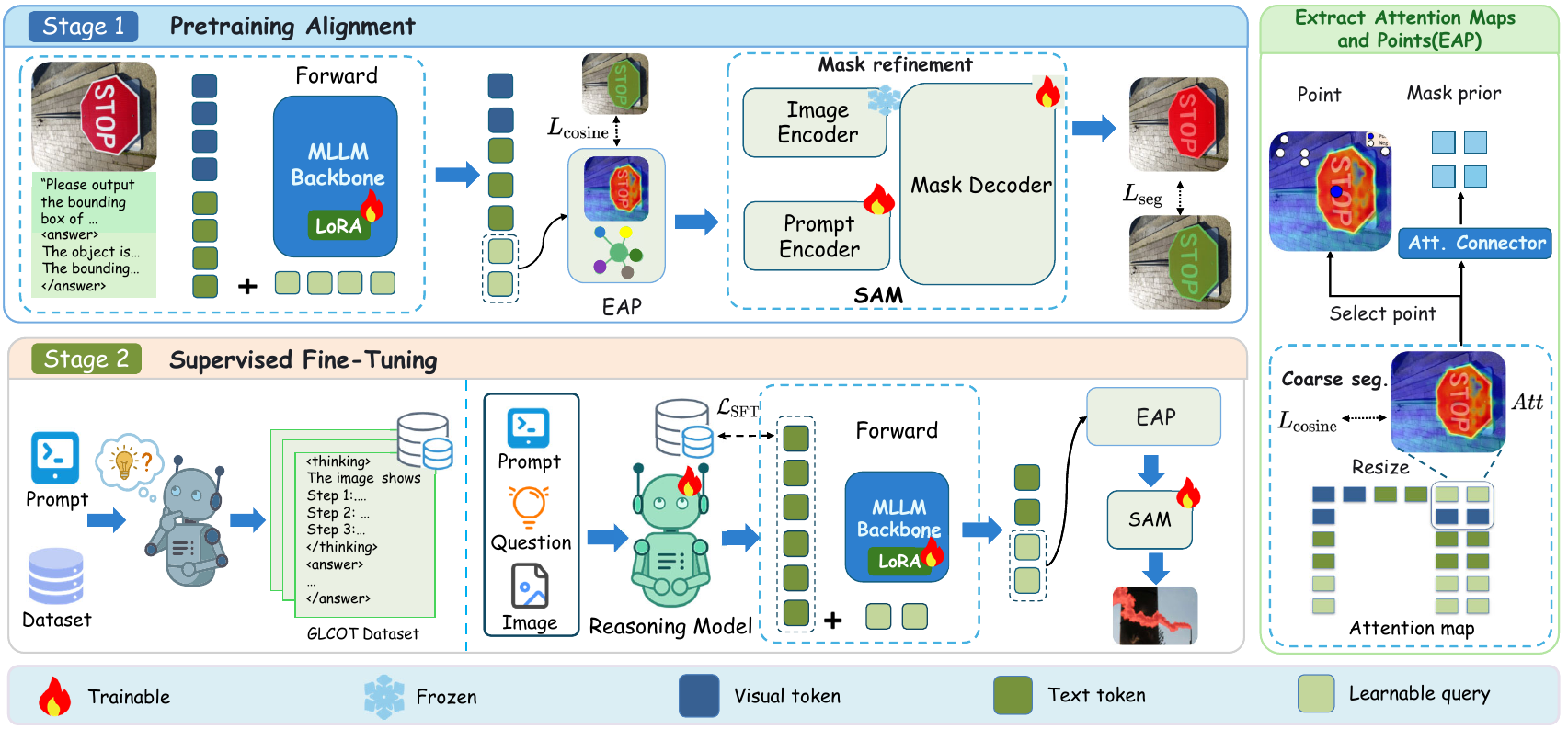}%
    }{%
      \framebox[\linewidth]{\rule{0pt}{180pt} figure1concept.pdf}%
    }
    \vspace{-13pt} 
    \caption{Overview of the two-stage training pipeline of \textsc{CR-Seg}. Stage 1: the image, question, and answer are fed jointly to train learnable query attention. Stage 2: the model is jointly trained on generation and segmentation using a GLCoT dataset distilled from a larger teacher model.Tokens enclosed by dashed boxes denote the tokens used for loss computation or subsequent processing. }
  \label{fig:3}
\end{figure*}

\section{Method}

To overcome the above limitations, we propose \textsc{CR-Seg}, an attention-guided coarse-to-refined reasoning segmentation framework with a two-stage training paradigm, as illustrated in Figure~\ref{fig:3}. Given an image and a question, the MLLM produces reasoning and answers, while learnable queries extract attention maps that are converted by the EAP module into SAM-compatible mask prior and points.

\subsection{\textsc{CR-Seg} Framework}
\textbf{Reasoning Model.} We use the open-source MLLM, Qwen3-VL-4B \citep{bai2025qwen3vltechnicalreport}, as our reasoning model.  Given an image $\mathcal{I}$, a prompt $\mathcal{P}$, and a question $\mathcal{Q}$, the MLLM generates a structured response, i.e.,
\texttt{<thinking>} $\mathcal{R}$ \texttt{</thinking>}
\texttt{<answer>} $\mathcal{A}$ \texttt{</answer>}, where $\mathcal{A}$ contains the predicted labels $\mathcal{A}_{\text{label}}$ and their corresponding boundary boxes $\mathcal{A}_{\text{box}}$.
\begin{equation}
  \label{eq:1}
  \mathcal{R},\mathcal{A} = MLLM(\mathcal{I},\mathcal{P},\mathcal{Q}).
\end{equation}

\noindent \textbf{Attention Extraction.} To extract the semantic information from the reasoning model’s outputs, we append learnable queries $\mathcal{L}_{Q} \in \mathbb{R}^{N \times D}$ to the end of the response, where $N$ is the number of learnable queries and $D$ is the hidden dimension of the MLLM. The concatenated sequence is then fed into the MLLM backbone for a forward pass.
\begin{equation}
  \label{eq:2}
  \mathcal{L}^{\prime}_{Q} = MLLM_{backbone}(\mathcal{I},\mathcal{P},\mathcal{Q},\mathcal{R},\mathcal{A},\mathcal{L}_{Q}).
\end{equation}
Here, $\mathcal{L}^{\prime}_{Q} \in \mathbb{R}^{N \times D}$ denotes the output embeddings of the learnable queries.
Afterward, the attention maps \citep{NIPS2017_3f5ee243} corresponding to the learnable queries are extracted. 
\noindent  However, with a large number of tokens, the Softmax normalization tends to produce overly flattened attention distributions, reducing the dynamic range of attention scores and weakening gradient signals propagated to SAM. To address this, we bypass the Softmax layer and directly employ the raw query-key affinity matrix as the attention map, thereby preserving cross-modal alignment while avoiding the information degradation caused by Softmax normalization. We denote the resulting attention map as $\operatorname{A}_{\mathcal{L}^{\prime}_{Q}} \in \mathbb{R}^{L \times K \times K}$, where $L$ is the number of MLLM layers and $K$ is the total number of tokens.
\begin{equation}
    \label{eq:4}
\operatorname{A}_{\mathcal{L}^{\prime}_{Q}}=\frac{Q K^{T}}{\sqrt{d}}.
\end{equation}

\noindent \textbf{Extract Attention Maps and Points (EAP).} The raw attention map $\operatorname{A}_{\mathcal{L}^{\prime}_{Q}}$ captures cross-modal relationships among all tokens, but it is not directly compatible with SAM's mask prompt format. To bridge this gap, we introduce the Extract Attention Maps and Points (EAP) module. As illustrated on the right of Figure~\ref{fig:3}, the EAP module transforms $\operatorname{A}_{\mathcal{L}^{\prime}_{Q}}$ into a mask prior that is spatially aligned with SAM's input space. We also select positive and negative points as explicit location cues, enabling the model to produce more precise segmentation results.

\textbf{1) Attention aggregation.} From the attention map $\operatorname{A}_{\mathcal{L}^{\prime}_{Q}}$, which
encodes both textual and visual information, we retain only the visual-token entries, yielding
$\operatorname{A}^{\prime}_{\mathcal{L}^{\prime}_{Q}} \in \mathbb{R}^{L \times V \times N}$, where $L$, $V$, and $N$
denote the number of Transformer layers, visual tokens, and learnable queries, respectively. To obtain a stable prior, we aggregate across the layer and query dimensions via averaging. We denote the resulting map as $\operatorname{Att} \in \mathbb{R}^{H \times W}$, which is then resized to SAM’s input resolution $(H, W)$, e.g., $H = W = 1008$ for SAM~3. The aggregated attention map $\operatorname{Att}$ is supervised by the ground-truth mask $M_{\text{gt}}$ as a coarse segmentation prior, which SAM later refines into a precise mask.

\textbf{2) Point sampling.} Let $\mathcal{S}$ denote all $H \times W$ spatial locations in $\operatorname{Att}$.
We construct two ordered sequences $\mathcal{S}_{\text{fg}}$ and $\mathcal{S}_{\text{bg}}$
by sorting $\mathcal{S}$ in descending and ascending order of attention values. We define the total attention mass as
\begin{equation}
  \label{eq:att_mass}
  M_{\mathrm{att}} = \sum_{i=1}^{H}\sum_{j=1}^{W} \sigma(\operatorname{Att}_{ij}),
\end{equation}
The foreground region $\mathcal{F}$ is the minimal prefix of
$\mathcal{S}_{\text{fg}}$ whose cumulative attention mass exceeds
$\alpha_{\mathrm{fg}} M_{\mathrm{att}}$.
The background region $\mathcal{B}$ is defined similarly from
$\mathcal{S}_{\text{bg}}$ with threshold $\alpha_{\mathrm{bg}}$.
 \begin{equation}
  \mathcal{F} = \min \left\{ \mathcal{S}_{\text{fg}} : \sum_{p \in \mathcal{S}_{\text{fg}}} \sigma(\operatorname{Att}_p) \ge \alpha_{\text{fg}} M_{\text{att}}
   \right\},
\end{equation}
\begin{equation}
  \mathcal{B} = \min \left\{ \mathcal{S}_{\text{bg}} : \sum_{p \in \mathcal{S}_{\text{bg}}} \sigma(\operatorname{Att}_p) \ge \alpha_{\text{bg}} M_{\text{att}}
   \right\}.
  \end{equation}
Where $\alpha_{\text{fg}}, \alpha_{\text{bg}}$ are preset hyperparameters.
From these regions, we construct the positive point set $P_{\text{pos}}$ by selecting the location with the maximum
attention in $\mathcal{F}$, plus four additional points sampled with probability proportional to their attention
values. The negative point set $P_{\text{neg}}$ consists of five points uniformly sampled from $\mathcal{B}$. 

\textbf{3) Mask prior.} Since the magnitude of $\operatorname{Att}$ differs from the feature scale expected by SAM, we apply a lightweight connector of four convolutional layers to project $\operatorname{Att}$ into SAM's input space, producing the mask prior $M_{P}$. The overall EAP module is summarized as:
\begin{equation}
  \label{eq:5}
  M_{P}, P_{\text{pos}}, P_{\text{neg}} = \operatorname{EAP}(\operatorname{A}_{\mathcal{L}^{\prime}_{Q}}).
\end{equation}

Further details are provided in Appendix~\ref{sec:EAP Module Workflow}.

\noindent \textbf{Segmentation Model.} SAM3 \citep{carion2025sam3segmentconcepts} is a mainstream segmentation model capable of generating precise segmentation masks conditioned on various prompts, including text, points, and bounding boxes. To achieve accurate segmentation, we feed $M_P$, $P_{\text{pos}}$, and $P_{\text{neg}}$ obtained from the EAP module into SAM as prompts. These prompts provide both mask prior and spatial information, while SAM is responsible for refining the segmentation boundaries to produce the final prediction $M_{\text{pred}}$.

\subsection{\textbf{Pretraining Alignment Stage}}
As shown at the top of Figure~\ref{fig:3}, this stage establishes the connection between the MLLM and SAM without optimizing the model's generative capability. To this end, we directly impose a loss on the attention map $\operatorname{Att}$, enabling rapid alignment between MLLM attention priors and segmentation masks. 
To further accelerate training, we apply LoRA \citep{hu2022lora} fine-tuning to the MLLM to facilitate faster adaptation of $\operatorname{Att}$. Meanwhile, to map $\operatorname{Att}$ into the input space of SAM, we unfreeze the Prompt Encoder and Mask Decoder of SAM.

To jointly optimize the alignment of the attention maps $\operatorname{Att}$ and the segmentation performance, we employ a cosine similarity loss $\mathcal{L}_{cos}$  \citep{8354191} and a segmentation loss $\mathcal{L}_{\text {seg }}$. The overall training objective $\mathcal{L}_{ali}$ is formulated as a weighted sum of these two terms, where the balance is controlled by $\lambda_{1}$ and $\lambda_{2}$:
\begin{equation}
    \label{eq:7}
    \mathcal{L}_{ali}=\lambda_{1} \mathcal{L}_{cos}+\lambda_{\text {2 }} \mathcal{L}_{\text {seg }}
\end{equation}
where $\mathcal{L}_{\text {seg }}$ denotes the segmentation loss, which consists of DICE loss \citep{7785132}, binary cross-entropy(BCE) loss, and Boundary loss :
\begin{equation}
\label{eq:8}
\begin{aligned}
\mathcal{L}_{\text{seg}} 
&= \lambda_{\text{dice}} \,\mathcal{L}_{\text{DICE}}(M_{\text{pred}}, M_{\text{gt}}) \\
&\quad + \lambda_{\text{bce}} \,\mathcal{L}_{\text{BCE}}(M_{\text{pred}}, M_{\text{gt}}) \\
&\quad + \lambda_{\text{bound}} \,\mathcal{L}_{\text{Boundary}}(M_{\text{pred}}, M_{\text{gt}}),
\end{aligned}
\end{equation}
where, following prior edge-aware methods \citep{DBLP:journals/corr/MathieuCL15,chen2019boundaryawarenetworkfasthighaccuracy}, we define a boundary loss by extracting Sobel edge maps and enforcing their L1 consistency between predictions and ground truth.
\begin{equation}
    \label{eq:9}
    K_x =
\begin{bmatrix}
1 & 0 & -1 \\
2 & 0 & -2 \\
1 & 0 & -1
\end{bmatrix},
\quad
K_y = K_x^{T},
\end{equation}

\begin{equation}
    \label{eq:10}
    E_{\text{pred}} = \left| M_{pred} * K_x \right| + \left| M_{pred} * K_y \right|,
\end{equation}

\begin{equation}
    \label{eq:11}
    E_{\text{gt}} = \left| M_{gt} * K_x \right| + \left| M_{gt} * K_y \right|,
\end{equation}

\begin{equation}
    \label{eq:12}
    \mathcal{L}_{\text{boundary}} = \frac{1}{HW} \sum_{i=1}^{H} \sum_{j=1}^{W} \left| E_{\text{pred}}^{(i,j)} - E_{\text{gt}}^{(i,j)} \right|.
\end{equation}

\subsection{\textbf{Supervised Fine-Tuning Stage}}
In this stage, we jointly train the model’s generative and segmentation capabilities. To reduce reasoning--answer inconsistency, we enforce an explicit three-step target localization process termed GLCoT, which begins with a global scene analysis, then progressively refines target localization at finer scales to produce the final answer.  
As illustrated in the lower-left part of Figure~\ref{fig:3}, we adopt a CoT distillation strategy \citep{hsieh-etal-2023-distilling,li-etal-2023-symbolic} to equip the model with this capability. A larger teacher model, Qwen3-VL-235B-A22B-Instruct, first generates GLCoT reasoning annotations, which are then stored in a GLCoT dataset. Further details are provided in Appendix \ref{sec:CoT data}.

We then use the GLCoT dataset to train the generation capability of the MLLM. Specifically, the SFT loss $\mathcal{L}_{\text{sft}}$ on GLCoT reasoning and the alignment loss $\mathcal{L}_{\text {ali}}$ on segmentation masks are optimized simultaneously, ensuring the model learns structured reasoning without sacrificing mask quality. The segmentation pipeline follows the same attention extraction and EAP procedure as in Stage~1. Since smaller target objects appear in this dataset, we restrict the positive point set $P_{\text{pos}}$ to a single point, i.e., the location with the highest attention value within the grounding box of the target object.
In this stage, we use the SFT loss and the alignment loss as the total loss.

\begin{equation}
    \label{eq:13}
    \begin{aligned}
    & \mathcal{L}=\lambda_{\text{sft}} \mathcal{L}_{SFT}+\lambda_{\text {ali}} \mathcal{L}_{\text {ali}}, \\
    & \mathcal{L}_{\mathrm{SFT}}=-\sum_{t=1}^{T} \log P_{\theta}\left(y_{t} \mid y_{<t}, x\right),
    \end{aligned}
\end{equation}
where $\mathcal{L}_{\text{ali}}$ represents the alignment loss introduced in the Pretraining Alignment Stage, and $\lambda_{\text{sft}}$ and $\lambda_{\text{ali}}$ are the corresponding balancing coefficients. In
$\mathcal{L}_{\mathrm{SFT}}$, $P_{\theta}$ denotes the model parameterized by $\theta$, $x$ is the input instruction, $y_t$ is the $t$-th token of the target response, $y_{<t}$ denotes the preceding tokens before position $t$, and $T$ is the total length of the response sequence.

\begin{table*}
\centering
\resizebox{0.95\textwidth}{!}{
\begin{tabular}{lc cc cc cc cc}
\toprule

\multirow{2}{*}{\textbf{Method}} 
& \multirow{2}{*}{\textbf{Train}} 
& \multicolumn{2}{c}{\textbf{L1}}
& \multicolumn{2}{c}{\textbf{L2}}
& \multicolumn{2}{c}{\textbf{L3}}
& \multicolumn{2}{c}{\textbf{Overall}} \\

\cmidrule(lr){3-10}

& & gIoU & cIoU 
& gIoU & cIoU 
& gIoU & cIoU 
& gIoU & cIoU \\

\midrule

\rowcolor{gray!15}
\multicolumn{10}{l}{\textit{Explicit-Prompt-Based Reasoning Segmentation (ERS)}} \\
Seg-Zero-7B \citep{liu2025segzeroreasoningchainguidedsegmentation}  & RL/Full FT 
& 59.22 & 47.38 
& 77.22 & 73.85 
& 77.86 & 75.03 
& 76.00 & 69.78 \\
VisionReasoner \citep{liu2026visionreasonerunifiedreasoningintegratedvisual} & RL/Full FT 
& 55.48 & 44.24 
& 72.92 & 68.93 
& 76.84 & 73.15 
& 73.29 & 66.23 \\

SAM3-Agent \citep{carion2025sam3segmentconcepts} & Agent
& 56.65 & 44.97 
& 64.47 & 63.45 
& 51.92 & 46.27 
& 57.86 & 54.49 \\

SAM3-Agent$^\dagger$ \citep{carion2025sam3segmentconcepts} & Agent
& 51.02 & 42.54 
& 72.96 & 68.54 
& 66.46 & 66.54 
& 68.02 & 63.34 \\

Evol-SAM3 \citep{ye2025evolvingtrainingzeroshotreasoning} & Agent
& 54.46 & 52.52
& 74.72 & 73.13 
& 68.05 & 64.69 
& 69.84 & 66.96 \\

Evol-SAM3$^\dagger$ \citep{ye2025evolvingtrainingzeroshotreasoning} & Agent
& 58.00 & 50.32 
& 65.43 & 60.22 
& 51.91 & 40.23 
& 58.40 & 50.44 \\

Dr.~Seg \citep{sun2026dr} &  RL/Full FT
& 55.64 & 45.46 
& 78.93 & 74.79 
& 78.01 & 76.94 
& 76.51 & 70.73 \\

Dr.~Seg$^\dagger$ \citep{sun2026dr} &  RL/LoRA
& \underline{61.04} & \textbf{58.74} 
& 76.37 & 73.53 
& \underline{80.06} & \underline{77.08} 
& 76.82 & 72.09 \\

\rowcolor{gray!15}
\multicolumn{10}{l}{\textit{Internal-Representation-Based Reasoning Segmentation (IRS)}} \\

LENS \citep{DBLP:conf/aaai/ZhuOZCHSRCYLW26} & RL/Full FT
& 59.80 & 51.51 
& \underline{79.05} & \underline{78.67} 
& 74.01 & 73.41 
& 75.03 & \underline{72.66} \\
LISA-7B \citep{lai2024lisa} & SFT/LoRA
& 44.1 & 48.17 
& 54.6 & 47.49 
& 70.33 & 70.98 
& 61.16 & 55.72 \\

GSVA \citep{Xia_2024_CVPR} & SFT/LoRA
& 59.25 & \underline{55.96} 
& 72.82 & 70.75 
& 63.44 & 55.10 
& 67.23 & 62.84 \\

\textbf{CR-Seg (ours)} & SFT/LoRA
& \textbf{61.55} & 51.69
& \textbf{82.70} & \textbf{79.74} 
& \textbf{80.48} & \textbf{78.54 }
& \textbf{79.85} & \textbf{74.56} \\

\bottomrule
\end{tabular}
}
\caption{\label{tab:Frasonseg}
    Performance on FReasonSeg, evaluated using gIoU and cIoU. \textbf{Bold} and \underline{underline} indicate the best and second-best performance, respectively. $\dagger$ denotes  results reproduced on Qwen3-VL-4B and SAM3.
  }
\end{table*}

\section{Experiments}

\subsection{Experiment Setup}

\textbf{Training data.} We use the RefCOCO series \citep{mao2016generation,yu2016modeling} as the training set for Stage~1. In Stage~2, we construct training set from the ReasonSeg \citep{lai2024lisa} training split, with GLCoT data generated by Qwen3-VL-235B-A22B-Instruct. 

\noindent \textbf{Benchmarks.} We adopt both ReasonSeg \citep{lai2024lisa} and our proposed FReasonSeg, which serves as a complementary benchmark. ReasonSeg heightens the difficulty of textual reasoning, yet most images lack visually similar distractors, allowing models to localize targets through textual reasoning without fine-grained visual discrimination. To fill this gap and evaluate the robustness of \textsc{CR-Seg} against attention interference from same-category objects, we construct FReasonSeg from existing RefCOCOm \citep{Wang_2024_CVPR} dataset. FReasonSeg is explicitly designed to stress-test robustness against same-category distractors in a zero-shot setting and comprises three progressively harder subsets—L1, L2, and L3—with L3 containing the highest proportion of same-category objects. Construction details are provided in Appendix \ref{sec:FReasonSeg}. 

\noindent \textbf{Evaluation Metrics.} For evaluation, following prior work \citep{liu2025segzeroreasoningchainguidedsegmentation,liu2026visionreasonerunifiedreasoningintegratedvisual,sun2026dr}, we report generalized IoU (gIoU), the per-image average IoU, and cumulative IoU (cIoU), the IoU between accumulated predictions and ground truths over the entire dataset.

\noindent \textbf{Implementation Details.} We adopt DeepSpeed \citep{rajbhandari2020zero} and conduct all experiments on 2 NVIDIA A6000 GPUs. For the Pretraining Alignment Stage, we set the batch size to 2, use a learning rate of 3e-5, and train for 2 epochs. For the SFT stage, we set the batch size to 1, use a learning rate of 1e-5, and train for 10 epochs. More detailed information is in Appendix \ref{sec:appendix More Implementation Detail}.

\subsection{Baselines}
We compare two categories of baselines: (1) \textbf{ERS methods}, including Seg-Zero \citep{liu2025segzeroreasoningchainguidedsegmentation}, VisionReasoner~\citep{liu2026visionreasonerunifiedreasoningintegratedvisual}, Dr.~Seg \citep{sun2026dr}, SAM3-Agent \citep{carion2025sam3segmentconcepts}, and Evol-SAM3 \citep{ye2025evolvingtrainingzeroshotreasoning}, where SAM3-Agent and Evol-SAM3 employ multi-step reasoning frameworks; and (2) \textbf{IRS methods}, including LISA \citep{lai2024lisa}, GSVA \citep{Xia_2024_CVPR}, and LENS \citep{DBLP:conf/aaai/ZhuOZCHSRCYLW26}.

\begin{figure}[t]
  \centering
    \IfFileExists{image/image_lq.pdf}{%
      \includegraphics[width=\linewidth]{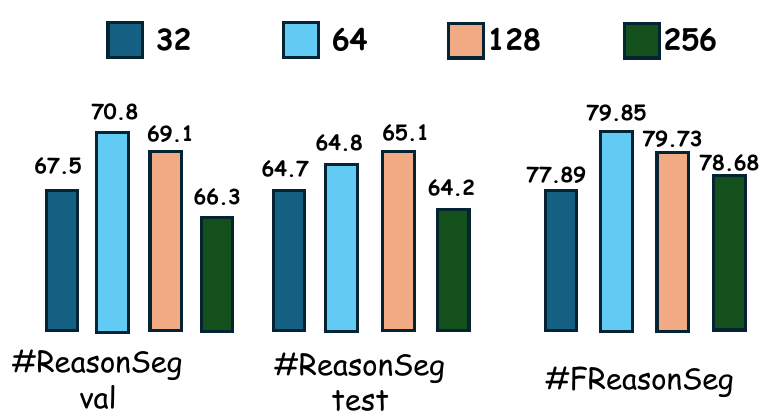}%
    }{%
      \framebox[\linewidth]{\rule{0pt}{180pt} image/image_lq.pdf}%
    }
    \caption{Comparison of different numbers of learnable queries on ReasonSeg and FReasonSeg.}
  \label{fig:lq}
\end{figure}

\begin{table*}
\centering
\resizebox{0.8\textwidth}{!}{
\begin{tabular}{l ccc ccc}
\toprule

\multirow{2}{*}{\textbf{Method}} 
& \multicolumn{2}{c}{\textbf{ReasonSeg-Val}} 
& \multicolumn{2}{c}{\textbf{ReasonSeg-Test}} 
& \multicolumn{2}{c}{\textbf{AVG}} \\

& gIoU & cIoU
& gIoU & cIoU
& gIoU & cIoU \\

\midrule

\rowcolor{gray!15}
\multicolumn{7}{l}{\textit{Explicit-Prompt-Based Reasoning Segmentation (ERS)}} \\
Seg-Zero-7B \citep{liu2025segzeroreasoningchainguidedsegmentation} 
& 62.6 & 62.0 & 57.5 & 52.0 & 60.1 & 57.0
 \\

RSVP \citep{lu-etal-2025-rsvp} 
& 64.7 & \underline{63.1} & 60.3 & \underline{60.0} & 62.5 & \underline{61.6}
 \\

VisionReasoner$^{*}$ \citep{liu2026visionreasonerunifiedreasoningintegratedvisual} 
& 65.8 & 55.0 & 62.8 & 55.1 & 64.2 & 55.1
 \\

Dr.~Seg$^{*}$ \citep{sun2026dr} 
& \underline{68.4} & 61.2 & \textbf{65.6} & 58.0 & \underline{67.0} & 59.6
 \\

Dr.~Seg$^\dagger$ \citep{sun2026dr} 
& 67.6 & 58.3 & 64.5 & 58.0 & 66.1 & 58.2
 \\

\rowcolor{gray!15}
\multicolumn{7}{l}{\textit{Internal-Representation-Based Reasoning Segmentation (IRS)}} \\

LISA-7B \citep{lai2024lisa} 
& 52.9 & 54.0 & 47.3& 48.4 & 50.1 & 51.2
 \\
 
HyperSeg \citep{Wei_2025_CVPR}
& 59.2 & 56.7 & - & - & - & -
\\

X-SAM \citep{DBLP:conf/aaai/WangQJHFZMLL26}
& 56.6 & 32.9 & 57.8 & 41.0 & 57.2 & 37
\\
LENS$^{*}$ \citep{DBLP:conf/aaai/ZhuOZCHSRCYLW26}
& 63.1 & 62.6 & 56.6 & 54.9 & 59.9 & 58.8
\\

\textbf{CR-Seg (ours)}
& \textbf{70.8} & \textbf{66.8} & \underline{64.8} & \textbf{62.6} & \textbf{67.8} & \textbf{64.7}
 \\

\bottomrule
\end{tabular}
}
\caption{\label{tab:Rasonseg}
    Performance on ReasonSeg. \textbf{Bold} and \underline{underline} indicate the best and second-best performance, respectively. $^{*}$ denotes results obtained from our independent experiments under the same experimental conditions. $-$ indicates unreported results. $\dagger$ denotes  results reproduced by us using LoRA fine-tuning on Qwen3-VL-4B and SAM3.
  }
\end{table*}

\subsection{Main Results}

As shown in Table~\ref{tab:Frasonseg}, on the FReasonSeg dataset, \textsc{CR-Seg} (ours) surpasses existing reasoning segmentation methods, achieving state-of-the-art performance. Notably, its strong performance on the L2 and L3 subsets demonstrates that the attention-based prior remains robust even under interference from same-category objects. This suggests that \textsc{CR-Seg} can effectively preserve target-specific spatial semantics even under interference from visually similar instances. In contrast, multi-step reasoning frameworks, such as SAM3-Agent and Evol-SAM3 (implemented with Qwen2.5-VL-7B and Qwen3-VL-4B), do not achieve comparable performance. We attribute this limitation to their stronger reliance on text-guided segmentation, which is more vulnerable to ambiguity in same-category scenarios. Although Evol-SAM3 additionally incorporates grounding outputs, its iterative optimization process primarily focuses on refining text prompts, limiting the benefits of multi-step reasoning for precise grounding-based segmentation.

\begin{table}
\centering
\resizebox{0.43\textwidth}{!}{
\begin{tabular}{lccc}
\toprule

\textbf{Method} & \textbf{Words} & \textbf{gIoU} $\uparrow$ & \textbf{RAI} $\downarrow$ \\

\midrule 

VisionReasoner
&79 & 65.8 & 11.5
 \\

Dr.~Seg
& 153 & 68.4 & 9.7
 \\

\textbf{CR-Seg$_{\text{w/o GLCoT}}$}
&93 & 69.1 & 8.2
 \\

\textbf{CR-Seg}
& 139 & \textbf{70.8} & \textbf{2.6}
 \\

\bottomrule
\end{tabular}
}
\caption{\label{tab:GLCoT}
    Ablation study of GLCoT on the ReasonSeg validation set. RAI refers to Reasoning–Answer Inconsistency.
  }
\end{table}

As shown in Table~\ref{tab:Rasonseg}, under identical experimental conditions, our model achieves competitive performance on ReasonSeg, surpassing the previous single-step state-of-the-art by 2.4\% on the validation set and attaining the best average results across validation and test splits. Notably, these results are attained with LoRA fine-tuning on a limited set of samples, while still maintaining competitive performance across both datasets. We attribute this performance to two key factors (1) attention maps serve as strong and stable priors for segmentation; (2) GLCoT mitigates the inconsistency between reasoning and final answers.

\begin{figure*}[t]
  \centering
    \IfFileExists{image/image2_1.pdf}{%
      \includegraphics[width=\linewidth]{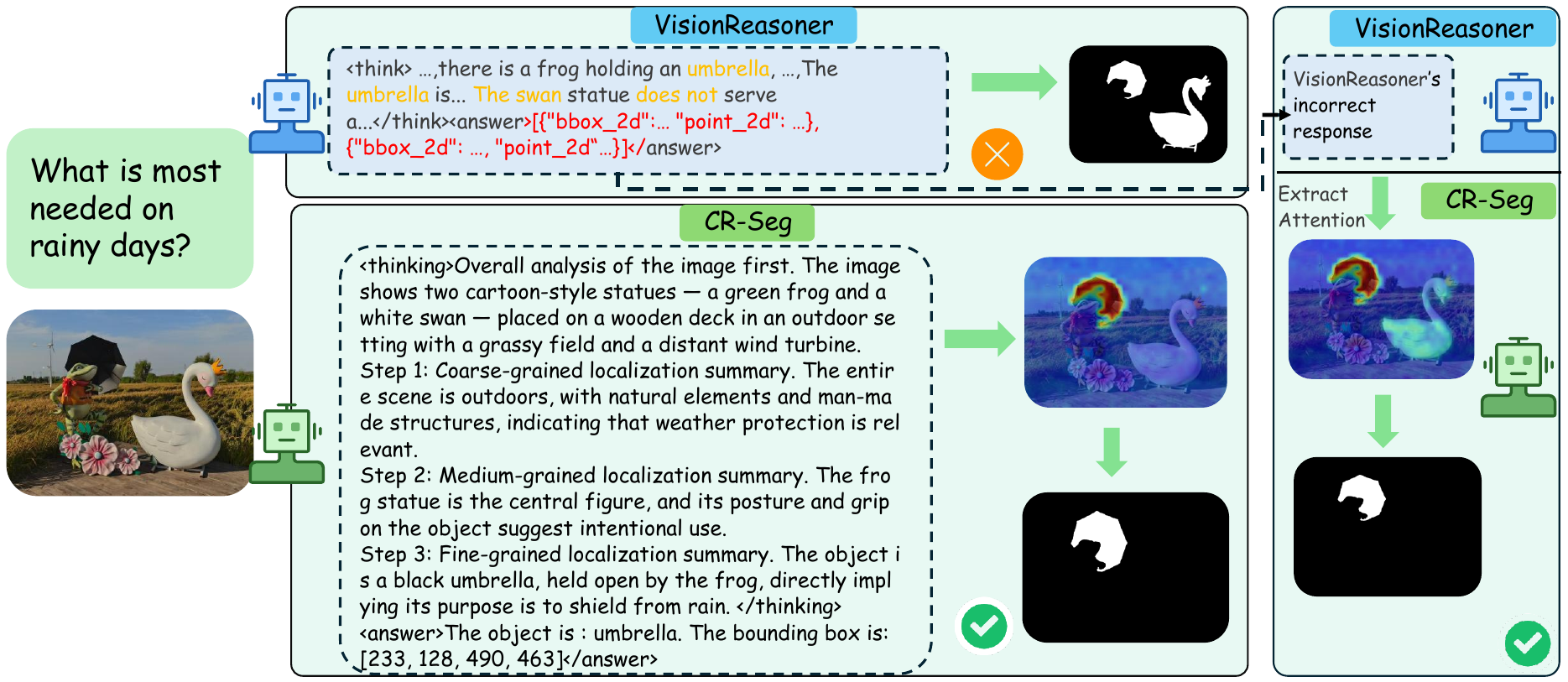}%
    }{%
      \framebox[\linewidth]{\rule{0pt}{180pt} figure1_concept.pdf}%
    }
    \vspace{-13pt} 
    \caption{Case study of a question where  \textsc{CR-Seg} answers correctly while VisionReasoner fails (Reasoning: umbrella; Answer: umbrella and swan).}
  \label{fig:2_1}
\end{figure*}

\subsection{Ablation Study}

\begin{table}
\centering
\resizebox{0.45\textwidth}{!}{
\begin{tabular}{c c c c c c}
\toprule

\multirow{2}{*}{Points} 
& \multicolumn{3}{c}{Pretraining Stage} 
& \multirow{2}{*}{\makecell{SFT \\ Stage}} 
& \multirow{2}{*}{gIoU} \\
\cmidrule(lr){2-4}
& MLLM & Conn. & SAM &  &  \\
\midrule

$\checkmark$ 
& $\checkmark$ & $\checkmark$ & $\checkmark$ & $\checkmark$ & 70.8 \\

$\times$ 
& $\checkmark$ & $\checkmark$ & $\checkmark$ & $\checkmark$ & 68.5 \\

$\checkmark$  
& $\times$ & $\checkmark$ & $\checkmark$ & $\checkmark$ & 67.4 \\

$\checkmark$  
& $\checkmark$ & $\checkmark$ & $\times$ & $\checkmark$ & 64.9 \\

$\checkmark$  
& - & - & - & $\checkmark$ & 55.9 \\

$\checkmark$  
& $\checkmark$ & $\checkmark$ & $\checkmark$ & - & 57.2 \\

\bottomrule
\end{tabular}
}
\caption{\label{tab:components}
Ablation study on the training stages and components of \textsc{CR-Seg} on ReasonSeg Val. $-$ denotes that the corresponding stage is skipped entirely.
}
\end{table}

\noindent \textbf{Ablation on the Number of Learnable Queries.} We investigate the impact of the number of learnable queries on model performance. As shown in Figure~\ref{fig:lq}, the model achieves optimal performance when the number of learnable queries is set to 64. However, as the number of learnable queries increases beyond this point, performance begins to degrade. We hypothesize that an excessive number of queries intensifies competition among them, which in turn hinders the model’s ability to effectively capture salient region information.

\textbf{Ablation on GLCoT.} To assess the contribution of GLCoT, we generate a variant CoT dataset containing only standard reasoning text (without three-step reasoning strategy) and retrain under the same SFT setting. We evaluate results on the ReasonSeg validation set using the modified CoT data. In addition, we adopt Qwen3.5-Plus \citep{qwen3.5} as the evaluator (every sample flagged as inconsistent was subsequently manually verified) and introduce a new metric, RAI (Reasoning–Answer Inconsistency),

\begin{equation}
    \label{eq:14}
    \mathrm{RAI}=\frac{N_{\text {inconsistent }}}{N_{\text {total }}}
\end{equation}
where $N_{\text{inconsistent}}$ counts samples whose reasoning contradicts the final answer. As shown in Table~\ref{tab:GLCoT}, without GLCoT, RAI is on par with the VisionReasoner and Dr.~Seg baselines. Incorporating GLCoT increases gIoU by 1.7\%, while reducing RAI by 5.6\%, demonstrating that GLCoT improves
\textsc{CR-Seg} performance while mitigating reasoning–answer inconsistency. Notably, \textsc{CR-Seg} does not produce the longest outputs among the compared methods, confirming that its lower RAI is not achieved through longer responses.

\noindent \textbf{Ablation on Components.} Table~\ref{tab:components} reports the ablation results on the ReasonSeg validation set. "Points" denotes the attention-map points extracted by the EAP module. MLLM, Conn., and SAM indicate whether each component is unfrozen during Stage~1, where Conn. refers to the connection module (learnable queries and the Att. Connector) between the MLLM and SAM. 
The results show that introducing points helps the model focus on key regions and improves overall performance. Both the pretraining alignment and SFT stages are important.  The pretraining alignment stage enables learnable queries to attend to salient regions and align attention maps with the SAM input space, while the SFT stage further encourages a global-to-local reasoning process, leading to significant performance gains.

\section{Case Study} 

We present a qualitative example in Figure~\ref{fig:2_1} to illustrate the robustness of \textsc{CR-Seg}. With GLCoT, \textsc{CR-Seg} maintains reasoning–answer consistency, correctly focusing on the target object despite the presence of distractors. To further stress-test this robustness, we provide the EAP module in \textsc{CR-Seg} with erroneous reasoning outputs generated by VisionReasoner, together with learnable queries. While this erroneous output leads VisionReasoner to produce an incorrect segmentation mask, \textsc{CR-Seg} still attends to the correct region and produces an accurate segmentation mask. This demonstrates that the attention-based bridge inherently captures correct visual semantics from the full MLLM response, making mask generation substantially more resilient to reasoning errors. More visualizations are provided in Appendix \ref{sec:Additional Visualizations}

\section{Conclusion}
We revisit reasoning segmentation and identify two key limitations in existing methods, where IRS methods struggle with semantic alignment while ERS methods fail to preserve holistic response semantics. To address both limitations, we propose \textsc{CR-Seg}, which uses attention maps as explicit and supervisable spatial priors to reduce the alignment difficulty of IRS methods with minimal alignment training overhead while preserving holistic semantics in MLLM outputs. Meanwhile, we introduce GLCoT to mitigate reasoning--answer inconsistency. Experiments show that \textsc{CR-Seg} achieves strong performance on reasoning segmentation tasks. Using MLLMs intrinsic capabilities more effectively for visual perception remains an open challenge. We hope this work serves as a useful foundation for future research in this direction.

\section*{Ethical Considerations}

Our work uses publicly available datasets and pretrained models strictly for academic research purposes. We ensure that all utilized data do not contain personally identifiable information or sensitive social information.
Potential risks mainly stem from biases inherited from pretrained MLLMs and possible misuse in surveillance-related applications.

\section*{Limitations}
Despite its effectiveness, \textsc{CR-Seg} remains dependent on SAM, and the attention maps supplied to SAM diverge from its native mask input format, limiting further gains. The framework also relies on the MLLM's internal attention mechanism and still requires additional training. Future work will therefore pursue two directions. One is to explore simpler refinement strategies for segmentation. The other is to move toward methods that require less additional training of the attention pathway.



\bibliography{main}

\begin{thebibliography}{39}
\providecommand{\natexlab}[1]{#1}

\bibitem[{Bai et~al.(2025{\natexlab{a}})Bai, Cai, Chen, Chen, Chen, Cheng, Deng, Ding, Gao, Ge, Ge, Guo, Huang, Huang, Huang, Hui, Jiang, Li, Li, Li, Li, Lin, Lin, Liu, Liu, Liu, Liu, Liu, Liu, Lu, Luo, Lv, Men, Meng, Ren, Ren, Song, Sun, Tang, Tu, Wan, Wang, Wang, Wang, Wang, Xie, Xu, Xu, Xu, Yang, Yang, Yang, Yang, Yu, Zhang, Zhang, Zhang, Zheng, Zhong, Zhou, Zhou, Zhou, Zhu, and Zhu}]{bai2025qwen3vltechnicalreport}
Shuai Bai, Yuxuan Cai, Ruizhe Chen, Keqin Chen, Xionghui Chen, Zesen Cheng, Lianghao Deng, Wei Ding, Chang Gao, Chunjiang Ge, Wenbin Ge, Zhifang Guo, Qidong Huang, Jie Huang, Fei Huang, Binyuan Hui, Shutong Jiang, Zhaohai Li, Mingsheng Li, and 45 others. 2025{\natexlab{a}}.
\newblock \href {https://arxiv.org/abs/2511.21631} {Qwen3-vl technical report}.
\newblock \emph{Preprint}, arXiv:2511.21631.

\bibitem[{Bai et~al.(2025{\natexlab{b}})Bai, Chen, Liu, Wang, Ge, Song, Dang, Wang, Wang, Tang, Zhong, Zhu, Yang, Li, Wan, Wang, Ding, Fu, Xu, Ye, Zhang, Xie, Cheng, Zhang, Yang, Xu, and Lin}]{bai2025qwen25vltechnicalreport}
Shuai Bai, Keqin Chen, Xuejing Liu, Jialin Wang, Wenbin Ge, Sibo Song, Kai Dang, Peng Wang, Shijie Wang, Jun Tang, Humen Zhong, Yuanzhi Zhu, Mingkun Yang, Zhaohai Li, Jianqiang Wan, Pengfei Wang, Wei Ding, Zheren Fu, Yiheng Xu, and 8 others. 2025{\natexlab{b}}.
\newblock \href {https://arxiv.org/abs/2502.13923} {Qwen2.5-vl technical report}.
\newblock \emph{Preprint}, arXiv:2502.13923.

\bibitem[{Carion et~al.(2025)Carion, Gustafson, Hu, Debnath, Hu, Suris, Ryali, Alwala, Khedr, Huang, Lei, Ma, Guo, Kalla, Marks, Greer, Wang, Sun, Rädle, Afouras, Mavroudi, Xu, Wu, Zhou, Momeni, Hazra, Ding, Vaze, Porcher, Li, Li, Kamath, Cheng, Dollár, Ravi, Saenko, Zhang, and Feichtenhofer}]{carion2025sam3segmentconcepts}
Nicolas Carion, Laura Gustafson, Yuan-Ting Hu, Shoubhik Debnath, Ronghang Hu, Didac Suris, Chaitanya Ryali, Kalyan~Vasudev Alwala, Haitham Khedr, Andrew Huang, Jie Lei, Tengyu Ma, Baishan Guo, Arpit Kalla, Markus Marks, Joseph Greer, Meng Wang, Peize Sun, Roman Rädle, and 19 others. 2025.
\newblock \href {https://arxiv.org/abs/2511.16719} {Sam 3: Segment anything with concepts}.
\newblock \emph{Preprint}, arXiv:2511.16719.

\bibitem[{Chen et~al.(2019)Chen, Qi, and Shen}]{chen2019boundaryawarenetworkfasthighaccuracy}
Xi~Chen, Donglian Qi, and Jianxin Shen. 2019.
\newblock \href {https://arxiv.org/abs/1901.03814} {Boundary-aware network for fast and high-accuracy portrait segmentation}.
\newblock \emph{Preprint}, arXiv:1901.03814.

\bibitem[{He et~al.(2026)He, Chen, Jian, Guo, Zhou, Li, Yang, and Yang}]{he2026dr2segdecomposedtwostagerollouts}
Yulin He, Wei Chen, Zhikang Jian, Tianhang Guo, Wenjuan Zhou, Minglong Li, Shaowu Yang, and Wenjing Yang. 2026.
\newblock \href {https://arxiv.org/abs/2601.09981} {Dr$^2$seg: Decomposed two-stage rollouts for efficient reasoning segmentation in multimodal large language models}.
\newblock \emph{Preprint}, arXiv:2601.09981.

\bibitem[{Hsieh et~al.(2023)Hsieh, Li, Yeh, Nakhost, Fujii, Ratner, Krishna, Lee, and Pfister}]{hsieh-etal-2023-distilling}
Cheng-Yu Hsieh, Chun-Liang Li, Chih-kuan Yeh, Hootan Nakhost, Yasuhisa Fujii, Alex Ratner, Ranjay Krishna, Chen-Yu Lee, and Tomas Pfister. 2023.
\newblock \href {https://doi.org/10.18653/v1/2023.findings-acl.507} {Distilling step-by-step! outperforming larger language models with less training data and smaller model sizes}.
\newblock In \emph{Findings of the Association for Computational Linguistics: ACL 2023}, pages 8003--8017, Toronto, Canada. Association for Computational Linguistics.

\bibitem[{Hu et~al.(2022)Hu, Shen, Wallis, Allen-Zhu, Li, Wang, Wang, and Chen}]{hu2022lora}
Edward~J Hu, Yelong Shen, Phillip Wallis, Zeyuan Allen-Zhu, Yuanzhi Li, Shean Wang, Lu~Wang, and Weizhu Chen. 2022.
\newblock \href {https://openreview.net/forum?id=nZeVKeeFYf9} {Lo{RA}: Low-rank adaptation of large language models}.
\newblock In \emph{International Conference on Learning Representations}.

\bibitem[{Kang et~al.(2025)Kang, Kim, Kim, and Hwang}]{Kang_2025_CVPR}
Seil Kang, Jinyeong Kim, Junhyeok Kim, and Seong~Jae Hwang. 2025.
\newblock Your large vision-language model only needs a few attention heads for visual grounding.
\newblock In \emph{Proceedings of the IEEE/CVF Conference on Computer Vision and Pattern Recognition (CVPR)}, pages 9339--9350.

\bibitem[{Kazemzadeh et~al.(2014)Kazemzadeh, Ordonez, Matten, and Berg}]{kazemzadeh2014referitgame}
Sahar Kazemzadeh, Vicente Ordonez, Mark Matten, and Tamara Berg. 2014.
\newblock Referitgame: Referring to objects in photographs of natural scenes.
\newblock In \emph{Proceedings of the 2014 conference on empirical methods in natural language processing (EMNLP)}, pages 787--798.

\bibitem[{Kirillov et~al.(2023)Kirillov, Mintun, Ravi, Mao, Rolland, Gustafson, Xiao, Whitehead, Berg, Lo, Dollar, and Girshick}]{Kirillov_2023_ICCV}
Alexander Kirillov, Eric Mintun, Nikhila Ravi, Hanzi Mao, Chloe Rolland, Laura Gustafson, Tete Xiao, Spencer Whitehead, Alexander~C. Berg, Wan-Yen Lo, Piotr Dollar, and Ross Girshick. 2023.
\newblock Segment anything.
\newblock In \emph{Proceedings of the IEEE/CVF International Conference on Computer Vision (ICCV)}, pages 4015--4026.

\bibitem[{Lai et~al.(2024)Lai, Tian, Chen, Li, Yuan, Liu, and Jia}]{lai2024lisa}
Xin Lai, Zhuotao Tian, Yukang Chen, Yanwei Li, Yuhui Yuan, Shu Liu, and Jiaya Jia. 2024.
\newblock Lisa: Reasoning segmentation via large language model.
\newblock In \emph{Proceedings of the IEEE/CVF conference on computer vision and pattern recognition}, pages 9579--9589.

\bibitem[{Li et~al.(2023)Li, Hessel, Yu, Ren, Chang, and Choi}]{li-etal-2023-symbolic}
Liunian~Harold Li, Jack Hessel, Youngjae Yu, Xiang Ren, Kai-Wei Chang, and Yejin Choi. 2023.
\newblock \href {https://doi.org/10.18653/v1/2023.acl-long.150} {Symbolic chain-of-thought distillation: Small models can also ``think'' step-by-step}.
\newblock In \emph{Proceedings of the 61st Annual Meeting of the Association for Computational Linguistics (Volume 1: Long Papers)}, pages 2665--2679, Toronto, Canada. Association for Computational Linguistics.

\bibitem[{Liu et~al.(2023)Liu, Li, Wu, and Lee}]{NEURIPS2023_6dcf277e}
Haotian Liu, Chunyuan Li, Qingyang Wu, and Yong~Jae Lee. 2023.
\newblock \href {https://proceedings.neurips.cc/paper_files/paper/2023/file/6dcf277ea32ce3288914faf369fe6de0-Paper-Conference.pdf} {Visual instruction tuning}.
\newblock In \emph{Advances in Neural Information Processing Systems}, volume~36, pages 34892--34916. Curran Associates, Inc.

\bibitem[{Liu et~al.(2025)Liu, Peng, Zhong, Yue, Lu, Yu, and Jia}]{liu2025segzeroreasoningchainguidedsegmentation}
Yuqi Liu, Bohao Peng, Zhisheng Zhong, Zihao Yue, Fanbin Lu, Bei Yu, and Jiaya Jia. 2025.
\newblock \href {https://arxiv.org/abs/2503.06520} {Seg-zero: Reasoning-chain guided segmentation via cognitive reinforcement}.
\newblock \emph{Preprint}, arXiv:2503.06520.

\bibitem[{Liu et~al.(2026)Liu, Qu, Zhong, Peng, Liu, Yu, and Jia}]{liu2026visionreasonerunifiedreasoningintegratedvisual}
Yuqi Liu, Tianyuan Qu, Zhisheng Zhong, Bohao Peng, Shu Liu, Bei Yu, and Jiaya Jia. 2026.
\newblock \href {https://arxiv.org/abs/2505.12081} {Visionreasoner: Unified reasoning-integrated visual perception via reinforcement learning}.
\newblock \emph{Preprint}, arXiv:2505.12081.

\bibitem[{Lu et~al.(2025)Lu, Cao, Wu, Li, Tang, Ji, Wu, Wu, and Zhu}]{lu-etal-2025-rsvp}
Yi~Lu, Jiawang Cao, Yongliang Wu, Bozheng Li, Licheng Tang, Yangguang Ji, Chong Wu, Jay Wu, and Wenbo Zhu. 2025.
\newblock \href {https://doi.org/10.18653/v1/2025.acl-long.715} {{RSVP}: Reasoning segmentation via visual prompting and multi-modal chain-of-thought}.
\newblock In \emph{Proceedings of the 63rd Annual Meeting of the Association for Computational Linguistics (Volume 1: Long Papers)}, pages 14699--14716, Vienna, Austria. Association for Computational Linguistics.

\bibitem[{Mao et~al.(2016)Mao, Huang, Toshev, Camburu, Yuille, and Murphy}]{mao2016generation}
Junhua Mao, Jonathan Huang, Alexander Toshev, Oana Camburu, Alan~L Yuille, and Kevin Murphy. 2016.
\newblock Generation and comprehension of unambiguous object descriptions.
\newblock In \emph{Proceedings of the IEEE conference on computer vision and pattern recognition}, pages 11--20.

\bibitem[{Mathieu et~al.(2016)Mathieu, Couprie, and LeCun}]{DBLP:journals/corr/MathieuCL15}
Micha{\"{e}}l Mathieu, Camille Couprie, and Yann LeCun. 2016.
\newblock \href {http://arxiv.org/abs/1511.05440} {Deep multi-scale video prediction beyond mean square error}.
\newblock In \emph{4th International Conference on Learning Representations, {ICLR} 2016, San Juan, Puerto Rico, May 2-4, 2016, Conference Track Proceedings}.

\bibitem[{Milletari et~al.(2016)Milletari, Navab, and Ahmadi}]{7785132}
Fausto Milletari, Nassir Navab, and Seyed-Ahmad Ahmadi. 2016.
\newblock \href {https://doi.org/10.1109/3DV.2016.79} {V-net: Fully convolutional neural networks for volumetric medical image segmentation}.
\newblock In \emph{2016 Fourth International Conference on 3D Vision (3DV)}, pages 565--571.

\bibitem[{Qian et~al.(2025)Qian, Yin, and Dou}]{qian2024reasoning}
Rui Qian, Xin Yin, and Dejing Dou. 2025.
\newblock Reasoning to attend: Try to understand how< seg> token works.
\newblock In \emph{Proceedings of the IEEE/CVF Conference on Computer Vision and Pattern Recognition}.

\bibitem[{{Qwen Team}(2026)}]{qwen3.5}
{Qwen Team}. 2026.
\newblock \href {https://qwen.ai/blog?id=qwen3.5} {{Qwen3.5}: Towards native multimodal agents}.

\bibitem[{Rajbhandari et~al.(2020)Rajbhandari, Rasley, Ruwase, and He}]{rajbhandari2020zero}
Samyam Rajbhandari, Jeff Rasley, Olatunji Ruwase, and Yuxiong He. 2020.
\newblock Zero: Memory optimizations toward training trillion parameter models.
\newblock In \emph{SC20: international conference for high performance computing, networking, storage and analysis}, pages 1--16. IEEE.

\bibitem[{Rasheed et~al.(2024)Rasheed, Maaz, Shaji, Shaker, Khan, Cholakkal, Anwer, Xing, Yang, and Khan}]{Rasheed_2024_CVPR}
Hanoona Rasheed, Muhammad Maaz, Sahal Shaji, Abdelrahman Shaker, Salman Khan, Hisham Cholakkal, Rao~M. Anwer, Eric Xing, Ming-Hsuan Yang, and Fahad~S. Khan. 2024.
\newblock Glamm: Pixel grounding large multimodal model.
\newblock In \emph{Proceedings of the IEEE/CVF Conference on Computer Vision and Pattern Recognition (CVPR)}, pages 13009--13018.

\bibitem[{Ravi et~al.(2025)Ravi, Gabeur, Hu, Hu, Ryali, Ma, Khedr, R{\"a}dle, Rolland, Gustafson et~al.}]{ravi2025sam}
Nikhila Ravi, Valentin Gabeur, Yuan-Ting Hu, Ronghang Hu, Chaitanya Ryali, Tengyu Ma, Haitham Khedr, Roman R{\"a}dle, Chloe Rolland, Laura Gustafson, and 1 others. 2025.
\newblock Sam 2: Segment anything in images and videos.
\newblock In \emph{International Conference on Learning Representations}, volume 2025, pages 28085--28128.

\bibitem[{Ren et~al.(2024)Ren, Huang, Wei, Zhao, Fu, Feng, and Jin}]{ren2024pixellm}
Zhongwei Ren, Zhicheng Huang, Yunchao Wei, Yao Zhao, Dongmei Fu, Jiashi Feng, and Xiaojie Jin. 2024.
\newblock Pixellm: Pixel reasoning with large multimodal model.
\newblock In \emph{Proceedings of the IEEE/CVF Conference on Computer Vision and Pattern Recognition}, pages 26374--26383.

\bibitem[{Sun et~al.(2026)Sun, Wang, Tang, Yuan, and Lv}]{sun2026dr}
Haoxiang Sun, Tao Wang, Chenwei Tang, Li~Yuan, and Jiancheng Lv. 2026.
\newblock Dr. seg: Revisiting grpo training for visual large language models through perception-oriented design.
\newblock \emph{arXiv preprint arXiv:2603.00152}.

\bibitem[{Vaswani et~al.(2017)Vaswani, Shazeer, Parmar, Uszkoreit, Jones, Gomez, Kaiser, and Polosukhin}]{NIPS2017_3f5ee243}
Ashish Vaswani, Noam Shazeer, Niki Parmar, Jakob Uszkoreit, Llion Jones, Aidan~N Gomez, \L~ukasz Kaiser, and Illia Polosukhin. 2017.
\newblock \href {https://proceedings.neurips.cc/paper_files/paper/2017/file/3f5ee243547dee91fbd053c1c4a845aa-Paper.pdf} {Attention is all you need}.
\newblock In \emph{Advances in Neural Information Processing Systems}, volume~30. Curran Associates, Inc.

\bibitem[{Wang et~al.(2026)Wang, Qiao, Jie, Huang, Feng, Zheng, Ma, Lan, and Liang}]{DBLP:conf/aaai/WangQJHFZMLL26}
Hao Wang, Limeng Qiao, Zequn Jie, Zhijian Huang, Chengjian Feng, Qingfang Zheng, Lin Ma, Xiangyuan Lan, and Xiaodan Liang. 2026.
\newblock \href {https://doi.org/10.1609/AAAI.V40I31.39822} {{X-SAM:} from segment anything to any segmentation}.
\newblock In \emph{Fortieth {AAAI} Conference on Artificial Intelligence, Thirty-Eighth Conference on Innovative Applications of Artificial Intelligence, Sixteenth Symposium on Educational Advances in Artificial Intelligence, {AAAI} 2026, Singapore, January 20-27, 2026}, pages 26187--26196. {AAAI} Press.

\bibitem[{Wang et~al.(2025)Wang, Fang, Kong, Li, Xu, Yang, Li, Zhu, and Wang}]{wang2025pixelthinkefficientchainofpixelreasoning}
Song Wang, Gongfan Fang, Lingdong Kong, Xiangtai Li, Jianyun Xu, Sheng Yang, Qiang Li, Jianke Zhu, and Xinchao Wang. 2025.
\newblock \href {https://arxiv.org/abs/2505.23727} {Pixelthink: Towards efficient chain-of-pixel reasoning}.
\newblock \emph{Preprint}, arXiv:2505.23727.

\bibitem[{Wang et~al.(2024)Wang, Yue, Zhang, Guo, He, Wang, and Liu}]{Wang_2024_CVPR}
Wenxuan Wang, Tongtian Yue, Yisi Zhang, Longteng Guo, Xingjian He, Xinlong Wang, and Jing Liu. 2024.
\newblock Unveiling parts beyond objects: Towards finer-granularity referring expression segmentation.
\newblock In \emph{Proceedings of the IEEE/CVF Conference on Computer Vision and Pattern Recognition (CVPR)}, pages 12998--13008.

\bibitem[{Wei et~al.(2025)Wei, Zhong, Tan, Liu, Hu, Li, Zhao, and Yang}]{Wei_2025_CVPR}
Cong Wei, Yujie Zhong, Haoxian Tan, Yong Liu, Jie Hu, Dengjie Li, Zheng Zhao, and Yujiu Yang. 2025.
\newblock Hyperseg: Hybrid segmentation assistant with fine-grained visual perceiver.
\newblock In \emph{Proceedings of the IEEE/CVF Conference on Computer Vision and Pattern Recognition (CVPR)}, pages 8931--8941.

\bibitem[{Wojke and Bewley(2018)}]{8354191}
Nicolai Wojke and Alex Bewley. 2018.
\newblock \href {https://doi.org/10.1109/WACV.2018.00087} {Deep cosine metric learning for person re-identification}.
\newblock In \emph{2018 IEEE Winter Conference on Applications of Computer Vision (WACV)}, pages 748--756.

\bibitem[{Wu et~al.(2024)Wu, Biamby, Chan, Dunlap, Gupta, Wang, Gonzalez, and Darrell}]{Wu_2024_CVPR}
Tsung-Han Wu, Giscard Biamby, David Chan, Lisa Dunlap, Ritwik Gupta, Xudong Wang, Joseph~E. Gonzalez, and Trevor Darrell. 2024.
\newblock See say and segment: Teaching lmms to overcome false premises.
\newblock In \emph{Proceedings of the IEEE/CVF Conference on Computer Vision and Pattern Recognition (CVPR)}, pages 13459--13469.

\bibitem[{Xia et~al.(2024)Xia, Han, Han, Pan, Song, and Huang}]{Xia_2024_CVPR}
Zhuofan Xia, Dongchen Han, Yizeng Han, Xuran Pan, Shiji Song, and Gao Huang. 2024.
\newblock Gsva: Generalized segmentation via multimodal large language models.
\newblock In \emph{Proceedings of the IEEE/CVF Conference on Computer Vision and Pattern Recognition (CVPR)}, pages 3858--3869.

\bibitem[{Ye et~al.(2025)Ye, You, Lin, Ji, Dai, and Cao}]{ye2025evolvingtrainingzeroshotreasoning}
Kai Ye, Xiaotong You, Jianghang Lin, Jiayi Ji, Pingyang Dai, and Liujuan Cao. 2025.
\newblock \href {https://arxiv.org/abs/2512.24702} {Evolving, not training: Zero-shot reasoning segmentation via evolutionary prompting}.
\newblock \emph{Preprint}, arXiv:2512.24702.

\bibitem[{Yu et~al.(2016)Yu, Poirson, Yang, Berg, and Berg}]{yu2016modeling}
Licheng Yu, Patrick Poirson, Shan Yang, Alexander~C Berg, and Tamara~L Berg. 2016.
\newblock Modeling context in referring expressions.
\newblock In \emph{European conference on computer vision}, pages 69--85. Springer.

\bibitem[{Zhang et~al.(2024)Zhang, Li, Fei, Yuan, Wu, Ji, Loy, and Yan}]{zhang2024omg}
Tao Zhang, Xiangtai Li, Hao Fei, Haobo Yuan, Shengqiong Wu, Shunping Ji, Chen~Change Loy, and Shuicheng Yan. 2024.
\newblock Omg-llava: Bridging image-level, object-level, pixel-level reasoning and understanding.
\newblock \emph{Advances in neural information processing systems}, 37:71737--71767.

\bibitem[{Zhou et~al.(2026)Zhou, Lai, Tan, Kil, Zhu, Chen, and Zhang}]{zhou2026guiaimaaligningintrinsicmultimodal}
Shijie Zhou, Viet~Dac Lai, Hao Tan, Jihyung Kil, Wanrong Zhu, Changyou Chen, and Ruiyi Zhang. 2026.
\newblock \href {https://arxiv.org/abs/2511.00810} {Gui-aima: Aligning intrinsic multimodal attention with a context anchor for gui grounding}.
\newblock \emph{Preprint}, arXiv:2511.00810.

\bibitem[{Zhu et~al.(2026)Zhu, Ouyang, Zhang, Cheng, Hu, Shen, Ran, Chen, Yu, Liu, and Wang}]{DBLP:conf/aaai/ZhuOZCHSRCYLW26}
Lianghui Zhu, Bin Ouyang, Yuxuan Zhang, Tianheng Cheng, Rui Hu, Haocheng Shen, Longjin Ran, Xiaoxin Chen, Li~Yu, Wenyu Liu, and Xinggang Wang. 2026.
\newblock \href {https://doi.org/10.1609/AAAI.V40I16.38405} {{LENS:} learning to segment anything with unified reinforced reasoning}.
\newblock In \emph{Fortieth {AAAI} Conference on Artificial Intelligence, Thirty-Eighth Conference on Innovative Applications of Artificial Intelligence, Sixteenth Symposium on Educational Advances in Artificial Intelligence, {AAAI} 2026, Singapore, January 20-27, 2026}, pages 13952--13960. {AAAI} Press.

\end{thebibliography}

\appendix

\section{Reasoning Segmentation Task}
\label{sec:Reasoning Segmentation Task}
Reasoning segmentation aims to segment target objects in an image according to complex language instructions that require visual-textual reasoning. Given an image and a reasoning-oriented query (e.g., “the equipment for sweeping away rain on rainy days.”), the model is required to identify the target object through compositional reasoning, and then output its corresponding segmentation mask. Compared with referring expression segmentation, where the language expression usually directly specifies the target object, reasoning segmentation involves more implicit and complex instructions that require the model to jointly understand both visual context and textual semantics to infer the correct target.

\section{Detailed Algorithm and Formulations}
\label{sec:Detailed Algorithm and Formulations}

\subsection{EAP Module Workflow}
\label{sec:EAP Module Workflow}

Algorithm \ref{alg:eap} outlines the EAP procedure. (i) In Steps~1--3, we extract the columns corresponding to the $N$ learnable queries from the attention map $\operatorname{A}_{\mathcal{L}^{\prime}_{Q}}$ and retain the rows associated with image patches. We then aggregate the $N$ query channels by averaging and resize the result to the resolution of the SAM input image $\mathcal{I}_{sam}$, yielding the final attention map $\operatorname{Att} \in \mathbb{R}^{H \times W}$. The attention map is used to generate two types of SAM inputs: a mask prior and point prompts. The mask prior $M_{P}$ is obtained by passing $\operatorname{Att}$ through a four-layer convolutional connector that preserves the input-output dimensionality. 
(ii) In Steps~4--6, we sort all values in $\operatorname{Att}$ in descending order, compute the total attention mass $M_{total}$, and compute the cumulative sum $C_k$ of the top-$k$ values. 
(iii) In Steps~7--10, we select the top-$k$ largest points to form the foreground region, where $k$ is the smallest value such that $C_k$ exceeds $\rho_{fp} \cdot M_{total}$ while ensuring that $k$ is no smaller than $N_{pos}$. The positive point set $P_{pos}$ includes the maximum-value point $v_{max}=\arg\max v$, and the remaining $N_{pos}-1$ points are sampled according to attention-based probabilities. 
(iv) The selection process for negative points is analogous. In Steps~11--14, a background region is determined based on the cumulative value $\rho_{bp} \cdot M_{total}$. Unlike positive points, negative points are not sampled according to attention probabilities; instead, $N_{neg}$ points are uniformly sampled from this background region.  
 
\begin{algorithm}[t]
\caption{Extract Attention Maps and Points (EAP)}
\label{alg:eap}
\begin{algorithmic}[1]

\Require 
Attention logits $\operatorname{A}_{\mathcal{L}'_{Q}} \in \mathbb{R}^{L \times K \times K}$; 
number of learnable queries $N$; 
number of image patches $I_p$; 
SAM image $\mathcal{I}_{sam} \in \mathbb{R}^{H \times W}$; 
connector network $Connector$; 
ratios $\rho_{fp}, \rho_{bp} \in (0,1)$; 
point numbers $N_{pos}, N_{neg}$

\Ensure 

\State $\operatorname{A}'_{\mathcal{L}'_Q} = \frac{1}{L} \sum_{l=1}^{L} \operatorname{A}_{\mathcal{L}'_Q}[l,:I_p,-N:]$

\State $Att \gets \text{Resize}(\operatorname{A}'_{\mathcal{L}'_Q}, (H,W))$

\State $M_P = Connector(Att)$

\State $\mathbf{p} \gets \operatorname{flatten}(\sigma(Att))$

\State $(\mathbf{v}, \boldsymbol{index}) \gets \operatorname{SortDesc}(\mathbf{p})$

\State $M_{total} \gets \sum_{t=1}^{H \times W} v_t$, \quad 
$C_k \gets \sum_{t=1}^{k} v_t$

\State $k_{fg} = \min\{k \mid C_k \ge \rho_{fp} \cdot M_{total}\}$

\State $k_{fg} = \max (k_{fg}, N_{pos})$

\State $\mathcal{I}_{fg} \gets \{\boldsymbol{index}_1, \ldots, \boldsymbol{index}_{k_{fg}}\}$

\State $P_{pos} = \{v_{max}\} \cup \exp\left(\frac{v_i}{\tau}\right), \; v_i \in \mathcal{I}_{fg}$

\State $k_{bg} = \min\{k \mid \sum_{i=1}^{k} v_{HW-i+1} \ge \rho_{bp} \cdot M_{total}\} $

\State $k_{bg} = \min (k_{bg},N_{neg})$

\State $\mathcal{I}_{bg} \gets \{\boldsymbol{index}_{HW-k_{bg}+1}, \ldots, \boldsymbol{index}_{HW}\}$

\State $P_{neg} \gets \text{RandomSample}(\mathcal{I}_{bg}, N_{neg})$

\State \Return $M_P, P_{pos}, P_{neg}$

\end{algorithmic}
\end{algorithm}

\section{Data Generation Process}
\label{sec:appendix Data Generation Process}

\subsection{FReasonSeg}
\label{sec:FReasonSeg}

\begin{figure*}[t]
  \centering
    \IfFileExists{image/image4.pdf}{%
      \includegraphics[width=\linewidth]{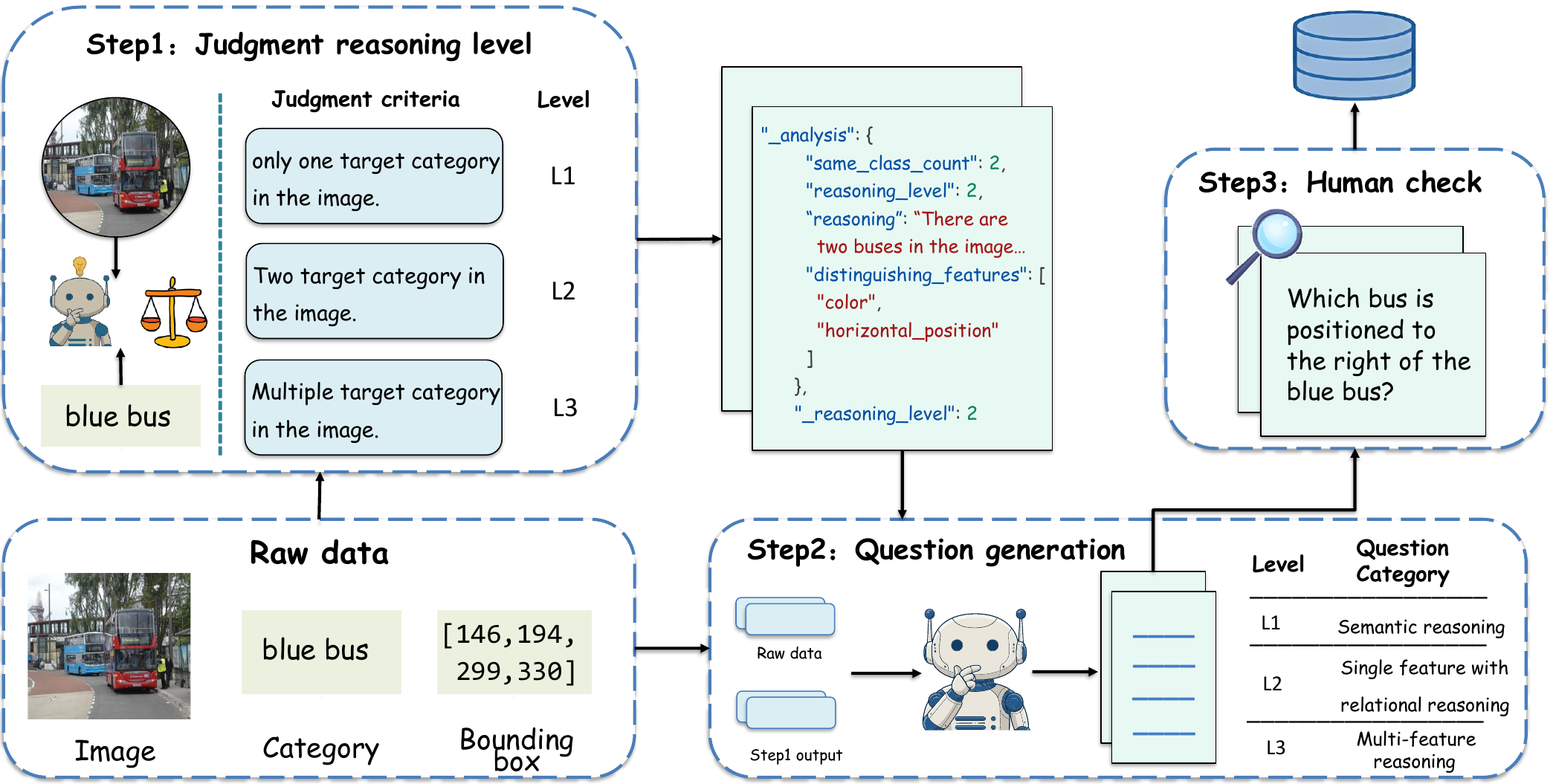}%
    }{%
      \framebox[\linewidth]{\rule{0pt}{180pt} figure1_concept.pdf}%
    }
    \vspace{-13pt} 
    \caption{Overview of the construction of FReasonSeg.}
  \label{fig:FReasonSeg}
\end{figure*}

Unlike ReasonSeg benchmarks, most images in FReasonSeg contain two or more visually similar target objects, requiring the model not only to reason but also to localize the correct instance based on fine-grained attribute descriptions, such as distance and size.

We use Qwen3-VL-235B-A22B-Instruct to construct the FReasonSeg dataset. We categorize the difficulty of each question based on the number of same-category objects in an image, and design different prompts accordingly to guide the model in generating questions. The detailed prompts are shown in Figures \ref{fig:reasoning_prompt}, \ref{fig:Generate Question Prompt(leval-1)}, \ref{fig:Generate Question Prompt(leval-2)}, and \ref{fig:Generate Question Prompt(leval-3)}.

Our construction process is divided into three stages, as shown in Figure~\ref{fig:FReasonSeg}. 

(1) \textbf{Reasoning-level annotation.} For each sample (image, category, target box) in RefCOCOm, we use Qwen3-VL-235B-A22B-Instruct to analyze the image and estimate the number of same-category objects. We then categorize reasoning difficulty into L1 (only one valid same-category object), L2 (two same-category objects, distinguishable by a single feature), and L3 (three or more same-category objects requiring multi-attribute composition).

(2) \textbf{Question generation.} We generate one question per sample conditioned on the assigned level, covering semantic reasoning for L1, single-feature relational discrimination for L2, and multi-feature compositional reasoning for L3.

(3) \textbf{Hunman check.} After data construction, we perform a multi-stage human check process to ensure the quality and usability of the dataset:

\begin{enumerate}
    \item \textbf{Candidate filtering:} We first remove all non-interrogative sentences. For the remaining samples, we further check whether each question is complete and well-formed.

    \item \textbf{Correctness filtering:} After the initial screening, we verify the consistency between each question and its corresponding label to prevent erroneous annotations from affecting evaluation results.

    \item \textbf{Quality inspection:} Finally, we assess whether each question satisfies the requirements of its designated difficulty level. Samples that fail to satisfy the intended reasoning difficulty are re-annotated to ensure consistency, producing the final benchmark.
\end{enumerate}

After filtering, the final FReasonSeg benchmark contains 283 samples in total, including L1 = 24, L2 = 125, and L3 = 134. The relatively small number of L1 samples is due to the fact that L1 does not contain same-category objects and therefore does not directly evaluate the model’s ability to distinguish visually similar instances. We include only a limited number of L1 cases for completeness and to maintain distributional coverage, while prioritizing L2 and L3, which focus on same-category discrimination.

\begin{figure}[t]
  \centering
    \IfFileExists{image/image5.pdf}{%
      \includegraphics[width=\linewidth]{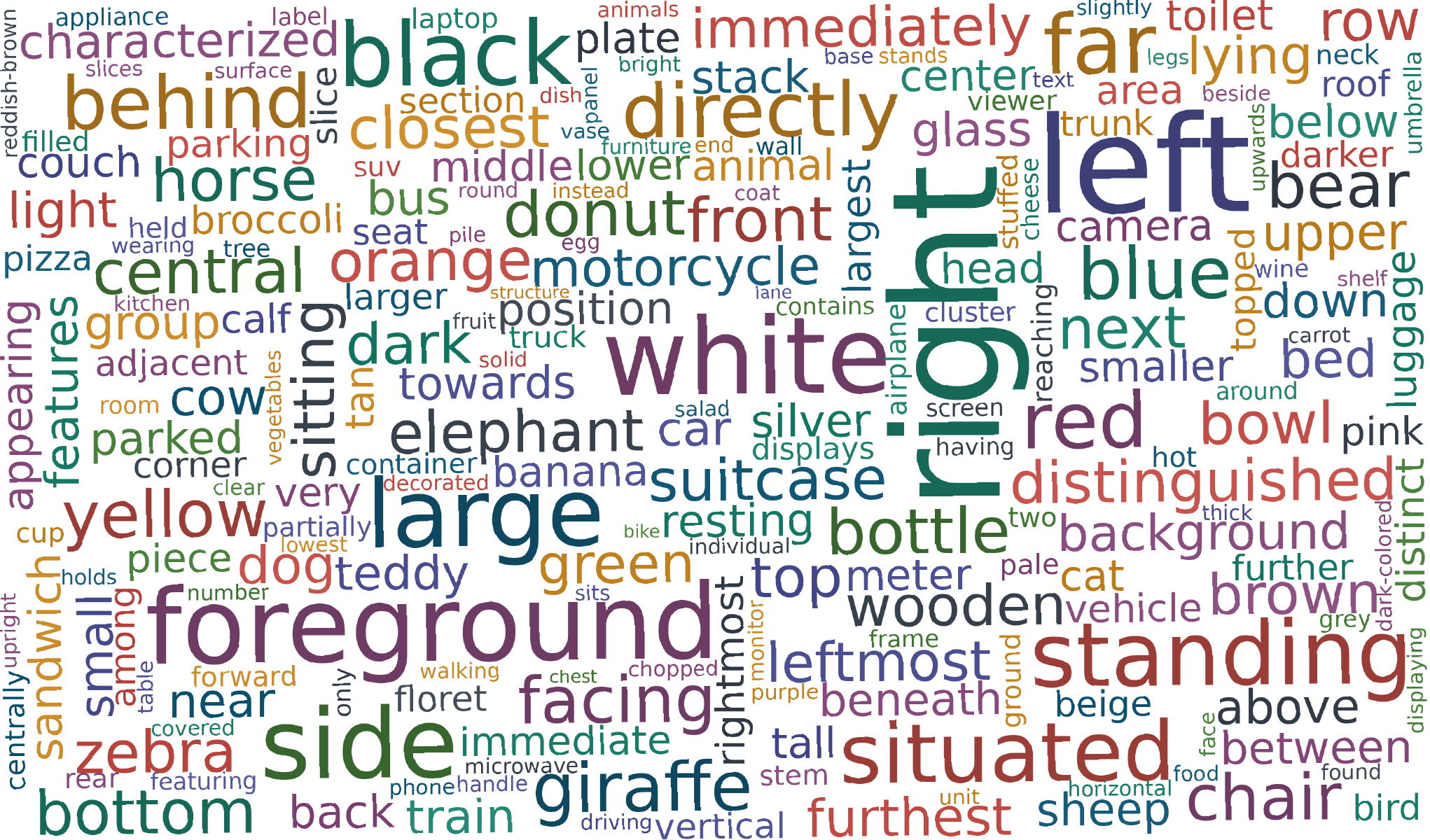}%
    }{%
      \framebox[\linewidth]{\rule{0pt}{180pt} image/image5.pdf}%
    }
    \caption{Word cloud visualization of the Top-220 most frequent words in FReasonSeg.}
  \label{fig:5}
\end{figure}

This benchmark evaluates the visual understanding ability of reasoning-based segmentation models. Given an image and a question, most images contain same-category objects, and the model is required to distinguish them based on attributes such as color and spatial location. Unlike ReasonSeg, which relies heavily on strong textual reasoning ability, this benchmark places greater emphasis on joint visual-language understanding before performing reasoning-based segmentation. We also provide overall statistics, as illustrated in the examples in Figure~\ref{fig:5}.

\begin{figure*}[t]
  \centering
    \IfFileExists{image/image6.pdf}{%
      \includegraphics[width=\linewidth]{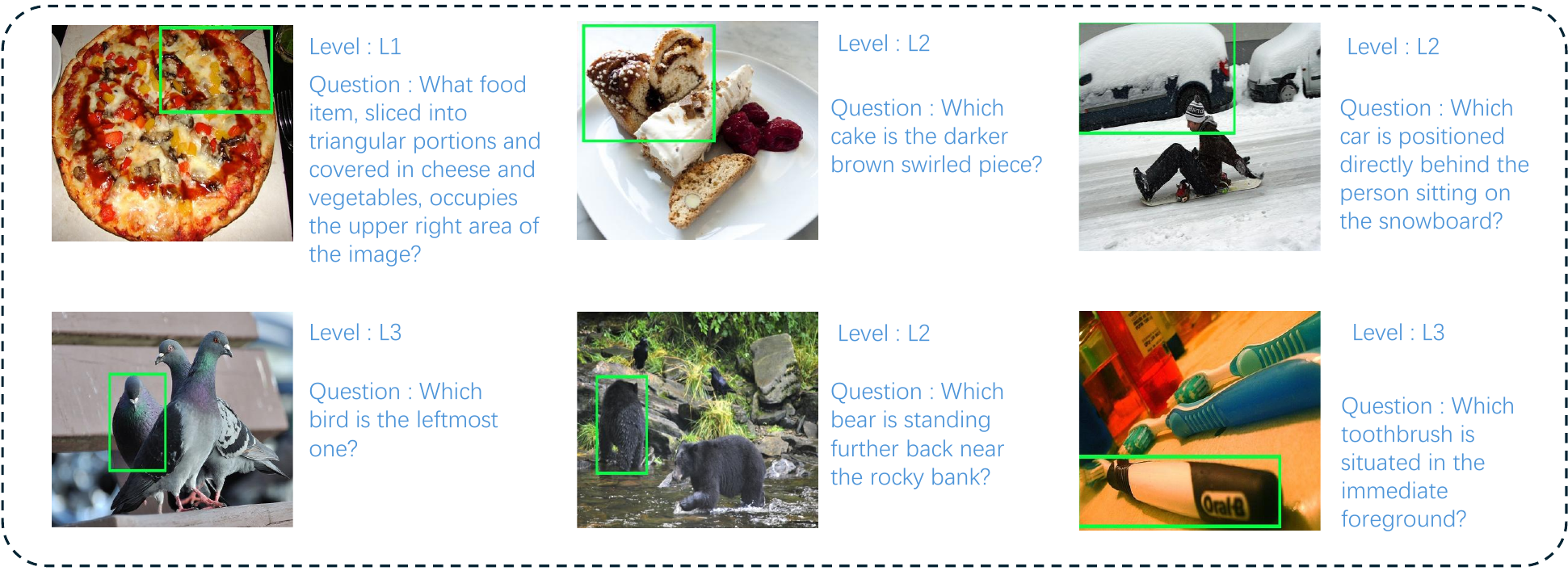}%
    }{%
      \framebox[\linewidth]{\rule{0pt}{180pt} image/image6.pdf}%
    }
    \caption{Visualization of examples from the FReasonSeg dataset.}
  \label{fig:6}
\end{figure*}

Furthermore, to provide a more intuitive illustration of the characteristics of our dataset, we visualize several examples from the FReasonSeg dataset, as shown in Figure~\ref{fig:6}.

\subsection{GLCoT data}
\label{sec:CoT data}

We construct our training data based on the ReasonSeg Train dataset, where Qwen3-VL-235B-A22B-Instruct is employed as the teacher model to generate reasoning chains. The detailed prompts used for data generation are presented in Figure~\ref{fig:cot_prompt_1} and \ref{fig:cot_prompt_2}. Since a single image in the ReasonSeg Train set may correspond to multiple questions, we select up to three questions for each image to increase the diversity and richness of the training data.

During the data generation process, each sample consists of an original image paired with a corresponding question. To further improve the accuracy and reliability of the generated reasoning chains, we additionally provide the teacher model with the segmentation mask overlaid on the original image, enabling the model to better perceive the target region and produce more precise reasoning trajectories.

\begin{table}[t]
\centering
\caption{Pretraining alignment stage experimental settings.}
\label{tab:pretrain_setting}

\begin{tabular}{ll}
\toprule
\textbf{config} &  \\
\midrule

epochs & 2 \\

batch size & 2 \\

gradient accumulation steps & 4 \\

learning rate & $3 \times 10^{-5}$ \\

scheduler & Cosine \\

deepspeed & Zero-2 \\

number of query & 64 \\

lora r & 16 \\

lora alpha & 32 \\

target modules & q, k, v, o \\

total training parameters & 16M \\

$\lambda_1$, $\lambda_2$ & 1, 1 \\

$\lambda_{\text{dice}}$, $\lambda_{\text{bce}}$, $\lambda_{\text{bound}}$ & 1, 2, 1 \\

\bottomrule
\end{tabular}

\end{table}

\begin{table}[t]
\centering
\caption{SFT stage experimental settings.}
\label{tab:sft_setting}

\begin{tabular}{ll}
\toprule
\textbf{config} &  \\
\midrule

epochs & 10 \\

batch size & 1 \\

gradient accumulation steps & 4 \\

learning rate & $1 \times 10^{-5}$ \\

scheduler & Cosine \\

deepspeed & Zero-2 \\

number of query & 64 \\

lora r & 16 \\

lora alpha & 32 \\

target modules & \makecell[l]{q, k, v, o, \\ gate, up, down} \\

total training parameters & 37M \\

$\lambda_{\text{sft}}$ & 1 \\

$\lambda_1$, $\lambda_2$ & 2, 1 \\

$\lambda_{\text{dice}}$, $\lambda_{\text{bce}}$, $\lambda_{\text{bound}}$ & 1, 2, 1 \\

\bottomrule
\end{tabular}

\end{table}

\section{More Implementation Details}
\label{sec:appendix More Implementation Detail}

All experiments were conducted on 2 NVIDIA A6000 GPUs. We adopted Qwen3-VL-4B as the reasoning model and SAM3 as the segmentation model. We employ DeepSpeed\cite{rajbhandari2020zero} to reduce GPU memory consumption during training. The total computational cost is approximately 56 GPU hours.

\begin{figure*}[t]
  \centering
    \IfFileExists{image/image7.pdf}{%
      \includegraphics[width=\linewidth]{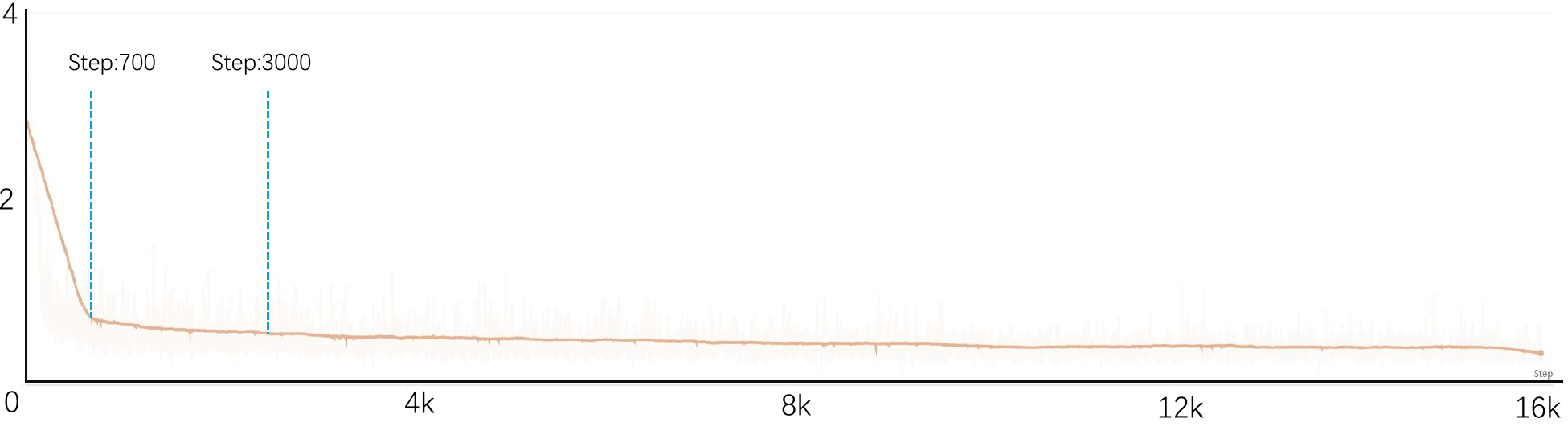}%
    }{%
      \framebox[\linewidth]{\rule{0pt}{180pt} image/image7.pdf}%
    }
    \caption{Loss dynamics throughout the pretraining alignment process.}
  \label{fig:7}
\end{figure*}

\noindent \textbf{Pretraining Alignment Stage.} 
The objective of this stage is to train the model to extract attention maps and align them with the input space of the segmentation model. The main experimental settings are summarized in Table~\ref{tab:pretrain_setting}. During this stage, the training data are derived from the RefCOCO series \citep{yu2016modeling, mao2016generation}. The loss curve over the entire training process is illustrated in Figure~\ref{fig:7}. Training already converges at 3k steps, achieving gIoU scores of 70.2 and 63.1 on ReasonSeg Val and Test, respectively. To obtain better overall performance, we still adopt the final checkpoint for second stage training.

\noindent \textbf{SFT Stage.} 
During this stage, we distill the reasoning trajectories (GLCoT) from Qwen3-VL-235B-A22B-Instruct to guide the model’s reasoning process. Detailed experimental configurations are presented in Table~\ref{tab:sft_setting}.

\noindent \textbf{Evaluation Details.} 
All experiments were conducted on 1 NVIDIA A6000 GPU. For FReasonSeg, all evaluated models use their publicly released checkpoints. For ReasonSeg, results marked with $^{*}$ are obtained by evaluating the publicly released checkpoints under the same experimental settings on all samples with available masks in the ReasonSeg val and test splits. For LENS, we adopt the version equipped with CoT reasoning.

\begin{table}\small
\centering
\normalsize
\resizebox{0.48\textwidth}{!}{
\begin{tabular}{cccc}
\toprule
\multirow{2}{*}{\textbf{Model}} & \multicolumn{2}{c}{\textbf{ReasonSeg}} & \multirow{2}{*}{\textbf{FReasonSeg}} \\
\cmidrule(lr){2-3}
 & Val & Test & \\
\midrule
\textbf{Qwen3-VL-4B+SAM3}       & 70.8 & 64.8 & 79.85 \\
Qwen3-VL-8B+SAM3    & 69.2 & 64.9 & \textbf{81.42} \\
Qwen3.5-4B+SAM3    & \textbf{72.0} & \textbf{68.0} & 76.30 \\
Qwen3-VL-4B+SAM2  & 70.8 & 65.3 & 79.23 \\
\bottomrule
\end{tabular}
}
\caption{Experiments on different model scaling and segmentation backbone.}
\label{tab:model}
\end{table}

\section{Additional Experiments}
\label{sec:More Eperiment}
As shown in Table~\ref{tab:model}, we scale the reasoning model from 4B to 8B. The results show improvements on two datasets, demonstrating the model’s scaling capability.

Substituting the base MLLM with Qwen3.5-4B \citep{qwen3.5}, which uses Hybrid Attention, further reveals that restricting attention extraction to only the eight full-attention layers produces more concentrated attention maps. This benefits ReasonSeg (where same-category discrimination is unnecessary) but degrades performance on FReasonSeg, which requires distinguishing visually similar objects.

Finally, replacing SAM3 with SAM2 results in a 0.62\% drop on FReasonSeg and a 0.5\% gain on ReasonSeg-Test. Overall, the impact remains relatively small.

\section{Additional Visualizations} 
\label{sec:Additional Visualizations}

\subsection{Additional Case Study}
\label{sec:Additional Case Study}

\begin{figure*}[t]
  \centering
    \IfFileExists{image/image8.pdf}{%
      \includegraphics[width=\linewidth]{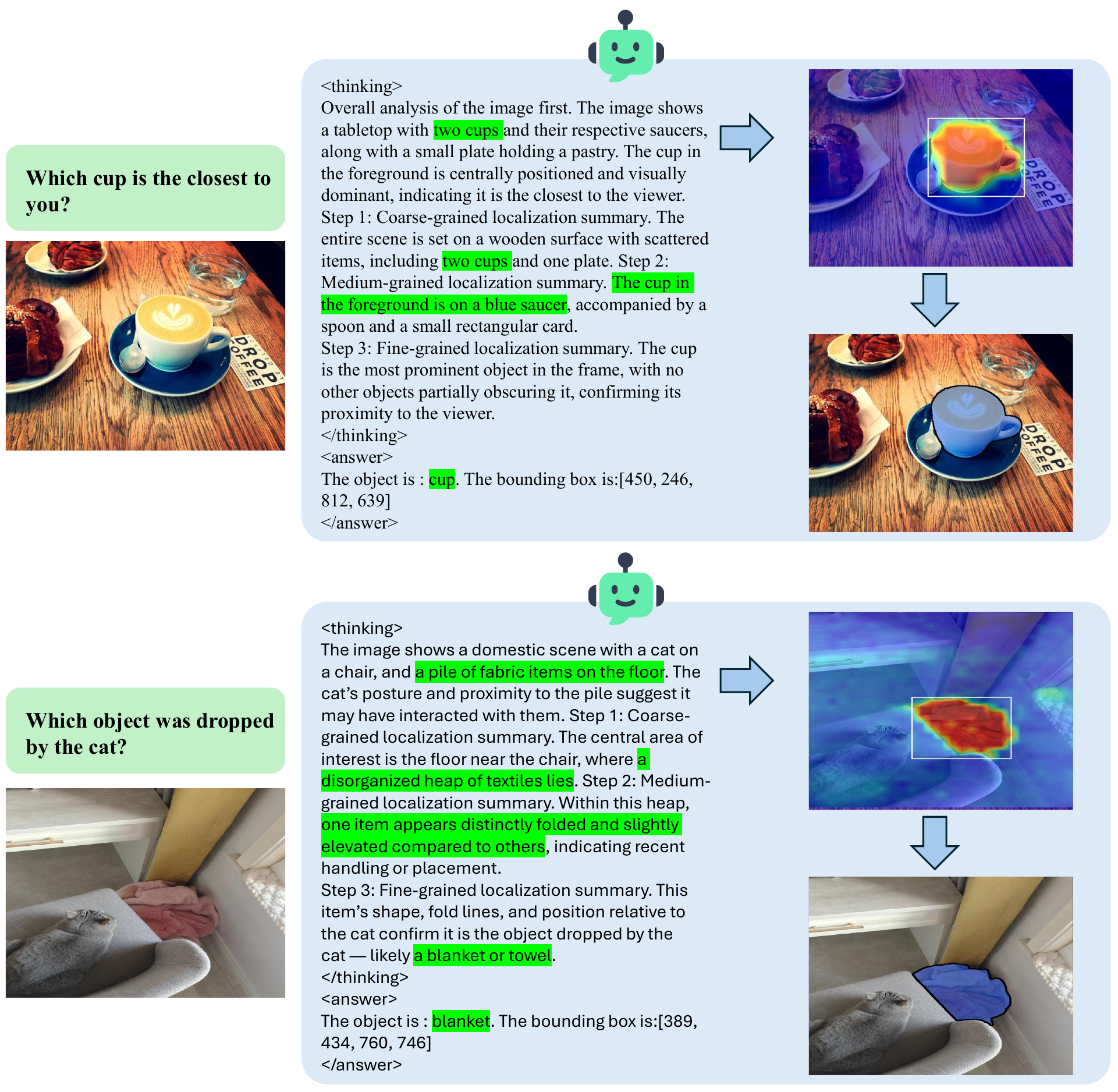}%
    }{%
      \framebox[\linewidth]{\rule{0pt}{180pt} image/image8.pdf}%
    }
    \caption{ Visualization of reasoning segmentation. The MLLM first generates a GLCoT, from which the learnable queries extract attention maps. After being processed by the EAP module, the segmentation model produces the final segmentation results.}
  \label{fig:8}
\end{figure*}

To further demonstrate the practical effectiveness of our framework in both reasoning and segmentation, we present two representative real-world examples in Figure~\ref{fig:8}. These cases highlight both the segmentation accuracy of \textsc{CR-Seg} and the coherence of its progressively refined reasoning process. The visualizations show that \textsc{CR-Seg} can perform fine-grained perceptual understanding and step-by-step reasoning, enabling accurate target localization under complex visual conditions.

\subsection{Progressive Attention Evolution in GLCoT}
\label{sec:Attention Evolution in GLCoT}

\begin{figure*}[t]
  \centering
    \IfFileExists{image/image_glcot_an.pdf}{%
      \includegraphics[width=\linewidth]{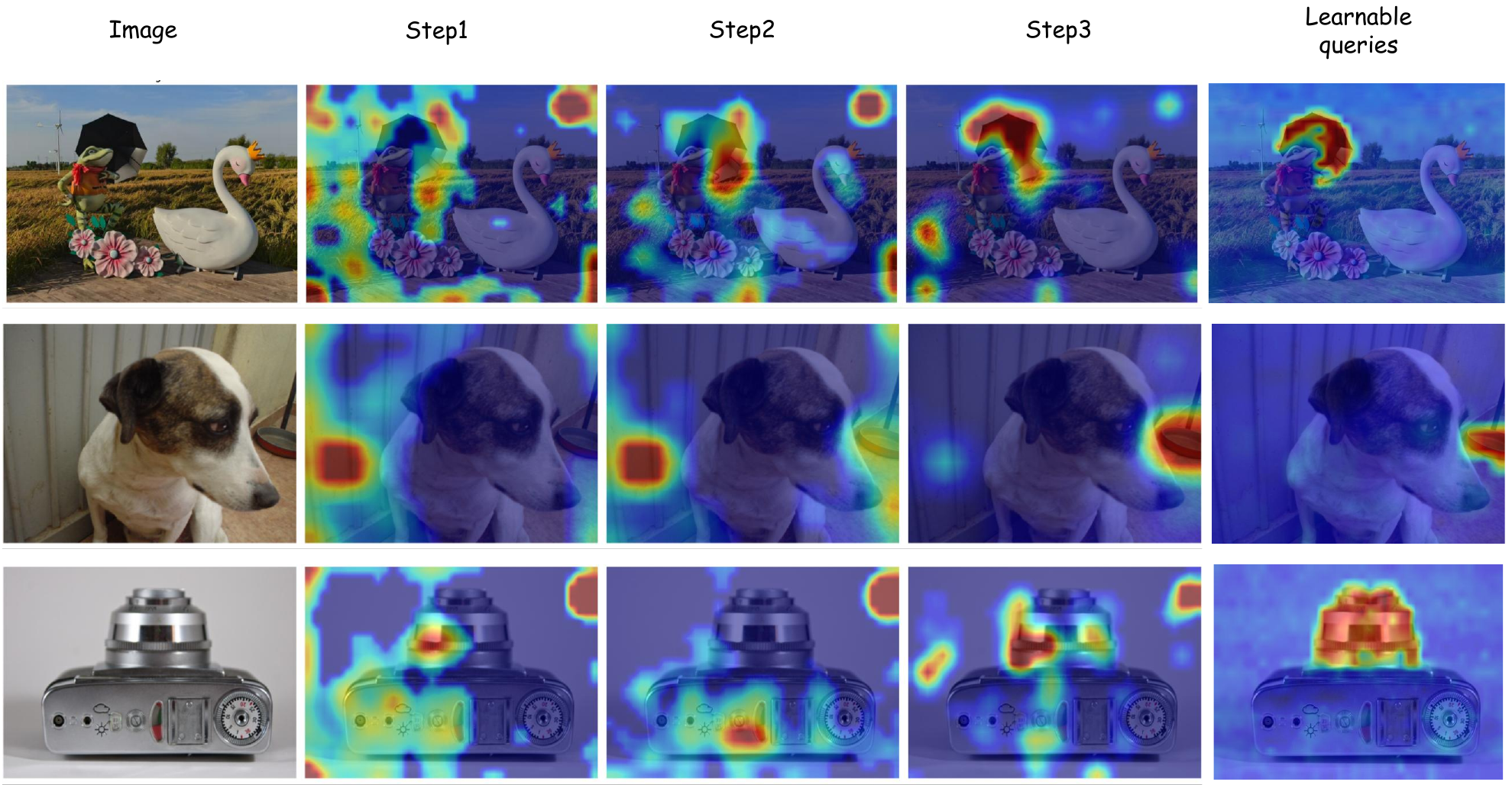}%
    }{%
      \framebox[\linewidth]{\rule{0pt}{180pt} image/image_glcot_an.pdf}%
    }
    \caption{Visualization of progressive attention refinement in GLCoT. We extract attention from the corresponding "step" tokens during reasoning, and from the learnable queries after the full response is generated.}
  \label{fig:glcot_an}
\end{figure*}
To verify that GLCoT drives progressive visual reasoning, we visualize the attention associated with the corresponding ``Step 1'', ``Step 2'', and ``Step 3'' tokens generated within the \texttt{\textless thinking\textgreater} reasoning block. As shown in Figure~\ref{fig:glcot_an}, attention systematically narrows from global scene regions in early steps to focused target boundaries in later steps, confirming that GLCoT induces a consistent coarse-to-fine visual grounding trajectory. This demonstrates that GLCoT guides the model to first observe the object globally and then progressively determine the final target.

\begin{figure*}[t]
\begin{tcolorbox}[
    width=\linewidth,
    colback=gray!10,        
    colframe=black!70,      
    fonttitle=\bfseries,
    arc=3mm,                
    boxrule=0.8pt,          
]

Please analyze this image, focusing on objects of category: \texttt{\{target\_category\}}.

The target object has bounding box coordinates (normalized to 0-1000): \texttt{\{normalized\_bbox\}}.

Please answer the following:

\begin{enumerate}[leftmargin=*, itemsep=1pt, topsep=1pt, parsep=0pt]
    \item How many \texttt{\{target\_category\}} objects are in the image?
    (count only complete, clearly visible ones)

    \item If only one, set
    \texttt{reasoning\_level} to \texttt{1}.

    \item If two, set
    \texttt{reasoning\_level} to \texttt{2},
    and describe what single feature can distinguish them
    (e.g., color, left/right position, size).

    \item If three or more, set
    \texttt{reasoning\_level} to \texttt{3},
    and describe what combination of features is needed
    to distinguish the target.
\end{enumerate}

\textbf{Output JSON format:}

\begin{verbatim}
{
    "same_class_count": number,
    "reasoning_level": 1/2/3,
    "reasoning": "explanation",
    "distinguishing_features": ["feature1", "feature2"]
}
\end{verbatim}

Output JSON only, no other text.
\end{tcolorbox}
\caption{Judgment reasoning level prompt}
\label{fig:reasoning_prompt}
\end{figure*}

\begin{figure*}[t]
\begin{tcolorbox}[
    width=\linewidth,
    colback=gray!10,        
    colframe=black!70,      
    fonttitle=\bfseries,
    arc=3mm,                
    boxrule=0.8pt,          
]

Generate a complex question for this target.

Target category: \texttt{\{category\_name\}}.

Target location: bounding box (normalized to 0-1000) \texttt{\{normalized\_bbox\}}

Requirements:

1. Must be a question ending with a question mark

2.The target object indicated by the question must be \texttt{\{category\_name\}}

3. Output only the question itself

Example: "What object in the picture can help humans eat?","Which part of the picture represents the waste discharged by the factory?"

Question:\texttt{\{question\}}

\end{tcolorbox}
\caption{Generate Question Prompt(leval-1)}
\label{fig:Generate Question Prompt(leval-1)}
\end{figure*}

\begin{figure*}[t]
\begin{tcolorbox}[
    width=\linewidth,
    colback=gray!10,        
    colframe=black!70,      
    fonttitle=\bfseries,
    arc=3mm,                
    boxrule=0.8pt,          
]

Generate a distinguishing question for this target.

Target category: \texttt{\{category\_name\}}.

Target location: bounding box (normalized to 0-1000)  \texttt{\{normalized\_bbox\}}

There are 2 \texttt{\{category\_name\}} objects in the image, need to distinguish using a single feature.
Available features: \texttt{\{features\_str\}}

Requirements:

1. Must be a question ending with a question mark

2. Use a single feature (e.g., left, red, large) to distinguish, but try to make it more complex(e.g., distance, relative to the position of other objects)

3. Output only the question itself

Example: "Which hot is on the left?","Which one is the hat closest to the person"

Question:\texttt{\{question\}}

\end{tcolorbox}
\caption{Generate Question Prompt(2)}
\label{fig:Generate Question Prompt(leval-2)}
\end{figure*}

\begin{figure*}[t]
\begin{tcolorbox}[
    width=\linewidth,
    colback=gray!10,        
    colframe=black!70,      
    fonttitle=\bfseries,
    arc=3mm,                
    boxrule=0.8pt,          
]

Generate a complex reasoning question for this target.

Target category: \texttt{\{category\_name\}}.

Target location: bounding box (normalized to 0-1000) \texttt{\{normalized\_bbox\}}

There are 3 or more \texttt{\{category\_name\}} objects in the image, need multiple features to distinguish.
Available features: \texttt{\{features\_str\}}

Requirements:

1. Must be a question ending with a question mark

2. Use multiple feature combinations (e.g., white one on the left, largest in the middle) to precisely locate

3. Description should require multi-step reasoning

4. Output only the question itself

Example: "What is the white hot on the left?"

Question:\texttt{\{question\}}

\end{tcolorbox}
\caption{Generate Question Prompt(leval-3)}
\label{fig:Generate Question Prompt(leval-3)}
\end{figure*}

\begin{figure*}[t]
\begin{tcolorbox}[
    width=\linewidth,
    colback=gray!10,
    colframe=black!70,
    arc=3mm,
    boxrule=0.8pt,
]

\small
\setlength{\parskip}{0.3em}

\texttt{
Note: I have highlighted the correct area in the image with a green overlay and green contour. Use this visual cue to guide your reasoning and accurately generate all bounding boxes. However, in your reasoning chain (COT and step\_cot), do NOT mention that you saw a highlighted region, green overlay, or green contour. Instead, describe your reasoning as if you identified the object purely from visual analysis.
}

\vspace{0.4em}

\texttt{Structure:}

\begin{enumerate}[leftmargin=1.2em,itemsep=0.15em,topsep=0.15em]
    \item First, provide an overall analysis of the entire image (NOT labeled as a step)
    \item Then provide Step 1: Coarse-grained localization -- summarize the general area
    \item Then provide Step 2: Medium-grained localization -- narrow down the region
    \item Then provide Step 3: Fine-grained localization -- pinpoint the exact object
    \item Finally, provide the answer with bounding boxes
\end{enumerate}

\vspace{0.2em}

\texttt{
IMPORTANT: The COT (Chain of Thought) must be concise, logical, and within 500 words. Focus on key observations and direct reasoning. Avoid unnecessary elaboration.
}

\vspace{0.4em}

\texttt{Example}

\vspace{0.2em}

\texttt{
Query: What is the object that the person in the picture is holding onto, while taking his or her dog for a walk?
}

\vspace{0.4em}

\begin{verbatim}
{
  "chain_of_thought":
  "Let's analyze... I notice... Step1: ... Step2: ... Step3: ... This is clearly a leash...",

  "reasoning_steps": [
    {
      "step": 1,
      "observation":"Person walking with dog in outdoor scene",
      "inference":"This is a dog walking scenario"
    },

    {
      "step": 2,
      "observation":"Linear object visible from hand to dog",
      "inference":"Most likely a leash given the context"
    },

    {
      "step": 3,
      "observation":"Object connects to dog's collar area",
      "inference":"Confirmed as leash -- a tether connecting handler to dog"
    }
  ],

  "final_answer": "leash"
}
\end{verbatim}

\vspace{0.4em}

\texttt{COT Output Format}

\vspace{0.2em}

\begin{verbatim}
<thinking>
The image shows a person walking a dog in an outdoor park setting. The person is holding something that
connects to the dog.
Step 1: Coarse-grained localization summary. The coarse scene shows a typical dog walking scenario 
in a park.
Step 2: Medium-grained localization summary. Focusing on the connection between the person's hand and 
the dogs reveals two linear objects.
Step 3: Fine-grained localization summary. This object extends from the hand to the two dogs' collar, 
clearly identifying it as two leashes used for walking the dog.
</thinking>
<answer>
The object is: leashes. The bounding box is: [43, 32, 178, 109], [352, 324, 747, 790]
</answer>
\end{verbatim}

\end{tcolorbox}

\caption{Generating GLCoT Data Prompt(1).}

\label{fig:cot_prompt_1}

\end{figure*}

\begin{figure*}[t]
\begin{tcolorbox}[
    width=\linewidth,
    colback=gray!10,
    colframe=black!70,
    arc=3mm,
    boxrule=0.8pt,
]

\small
\setlength{\parskip}{0.3em}

\texttt{
Note: In the example above, there are two objects related to the problem. If there is only one object, the following format should be output:
}

\begin{verbatim}
<answer>
The object is: object_name.
The bounding box is:
[x_min, y_min, x_max, y_max]
</answer>
\end{verbatim}

\vspace{0.4em}

\texttt{Required JSON Format}

\vspace{0.2em}

\begin{verbatim}
{
  "chain_of_thought":
  "Overall image analysis.
   Step 1: Coarse summary.
   Step 2: Medium summary.
   Step 3: Fine summary.",

  "reasoning_steps": [
    {
      "step": 1,
      "observation": "...",
      "inference": "..."
    },

    {
      "step": 2,
      "observation": "...",
      "inference": "..."
    },

    {
      "step": 3,
      "observation": "...",
      "inference": "..."
    }
  ],

  "final_answer": "object_name",

  "cot":
  "<thinking>\n
   Overall analysis of the image first. Then Step 1: Coarse-grained localization summary.
   Step 2: Medium-grained localization summary.
   Step 3: Fine-grained localization summary.\n
   </thinking>\n
   <answer>\n
   The object is: object_name.The bounding box is: [x_min, y_min, x_max, y_max]\n
   </answer>"
}
\end{verbatim}

\vspace{0.2em}

\texttt{REMEMBER:}

\begin{enumerate}[leftmargin=1.2em,itemsep=0.15em,topsep=0.15em]
    \item First analyze the entire image (overall context, not a step)
    \item Then provide Step 1, Step 2, and Step 3 as summaries of the reasoning\_steps
    \item Keep the COT concise (under 500 words), logical, and focused on key observations only
    \item Output the bounding boxes of all objects that satisfy the problem as required
    \item CRITICAL: The COT and step\_cot MUST NOT mention any \'highlighted region\', \'marked area\' or similar phrases. Describe your reasoning purely based on visual analysis of the image.
\end{enumerate}

\end{tcolorbox}

\caption{Generating GLCoT Data Prompt(2).}

\label{fig:cot_prompt_2}

\end{figure*}

\end{document}